\documentclass[a4paper,fleqn]{cas-dc}

\usepackage[numbers]{natbib}
\usepackage{lineno}
\def\tsc#1{\csdef{#1}{\textsc{\lowercase{#1}}\xspace}}
\tsc{WGM}
\tsc{QE}
\tsc{EP}
\tsc{PMS}
\tsc{BEC}
\tsc{DE}

\usepackage{amsmath,amsfonts}
\usepackage{algorithmic}
\usepackage{algorithm}
\usepackage{array}
\usepackage[caption=false,font=normalsize,labelfont=sf,textfont=sf]{subfig}
\usepackage{textcomp}
\usepackage{stfloats}
\usepackage{url}
\usepackage{verbatim}
\usepackage{graphicx}
\usepackage{cite}
\usepackage{soul}
\usepackage{booktabs}
\usepackage{pifont}
\usepackage{multicol}
\usepackage{tabularx}
\usepackage{multirow}
\usepackage{xcolor}
\usepackage{listings}
\usepackage{xspace}
\lstdefinelanguage{PDDL}{
    morekeywords={define, domain, requirements, types, predicates, action, parameters, precondition, effect, and, not},
    sensitive=true,
    morecomment=[l]{;},
    morestring=[b]"
}

\begin{document}
\let\WriteBookmarks\relax
\def\floatpagepagefraction{1}
\def\textpagefraction{.001}

\shorttitle{A Review of Generative AI in Smart Aquaculture}

\shortauthors{W. Akram et~al.}

\title[mode=title]{A Review of Generative AI in Aquaculture: Foundations, Applications, and Future Directions for Smart and Sustainable Farming}

\author[1]{Waseem Akram}
    \address[1]{Khalifa University Center for Autonomous Robotic Systems (KUCARS), Khalifa University, United Arab Emirates.}
    \ead{waseem.akram@ku.ac.ae}
    \credit{Conceptualization, Methodology, Software, Writing - Original Draft, Formal Analysis, Writing - Reviews and Editing}

 \author[1]{Muhayy Ud Din}
     \credit{Methodology, Software - Original Draft, Investigation }
     \ead{muhayyuddin.ahmed@ku.ac.ae}
     
 \author[1]{Lyes {Saad Saoud}}
     \credit{Methodology, Software - Original Draft, Investigation }
     \ead{lyes.saoud@ku.ac.ae}


\author[1]{Irfan Hussain}
     \credit{Methodology, Writing - Original Draft, Investigation }
     \cortext[cor1]{Corresponding author: I. Hussain (irfan.hussain@ku.ac.ae)}
\begin{abstract}
Generative Artificial Intelligence (GAI) has rapidly emerged as a transformative force in aquaculture, enabling intelligent synthesis of multimodal data, including text, images, audio, and simulation outputs for smarter, more adaptive decision-making. As the aquaculture industry shifts toward data-driven, automation and digital integration operations under the Aquaculture 4.0 paradigm, GAI models offer novel opportunities across environmental monitoring, robotics, disease diagnostics, infrastructure planning, reporting, and market analysis. This review presents the first comprehensive synthesis of GAI applications in aquaculture, encompassing foundational architectures (e.g., diffusion models, transformers, and retrieval augmented generation), experimental systems, pilot deployments, and real-world use cases. We highlight GAI's growing role in enabling underwater perception, digital twin modeling, and autonomous planning for remotely operated vehicle (ROV) missions. We also provide an updated application taxonomy that spans sensing, control, optimization, communication, and regulatory compliance.
Beyond technical capabilities, we analyze key limitations, including limited data availability, real-time performance constraints, trust and explainability, environmental costs, and regulatory uncertainty. We also identify pathways for overcoming these through multimodal modeling, federated learning, and domain-specific pretraining. Additionally, we provide an in-depth case study on the integration of GAI in marine robotics and control systems and propose future research directions for ethical, scalable, and sustainable GAI adoption. Our contributions include a structured mapping of GAI techniques to aquaculture tasks, a critical analysis of enabling technologies and barriers, and a roadmap toward responsible innovation. This review positions GAI not merely as a tool but as a critical enabler of smart, resilient, and environmentally aligned aquaculture systems.
\end{abstract}


\maketitle

\section{Introduction}

The global aquaculture industry, often referred to as the “blue revolution,” has undergone rapid expansion over recent decades, becoming a critical pillar for food security, nutrition, and economic development \citep{subasinghe2009global}. As the demand for aquatic food continues to rise, aquaculture systems must evolve beyond traditional methods toward more intelligent, resilient, and sustainable paradigms \citep{FAO2022}. Smart aquaculture or Aquaculture 4.0 uses Internet of Things (IoT) technologies, data analytics, robotics, and artificial intelligence (AI) to enhance productivity, mitigate environmental impact, and support real-time, data-driven decision-making \citep{Misra2020iott}. Nevertheless, the sector continues to grapple with complex challenges such as disease outbreaks, biofouling, labor-intensive operations, and operational risks in dynamic and harsh marine environments \citep{Shah2021smartt}.

\begin{figure}[t]
    \centering
    \includegraphics[width=1\linewidth]{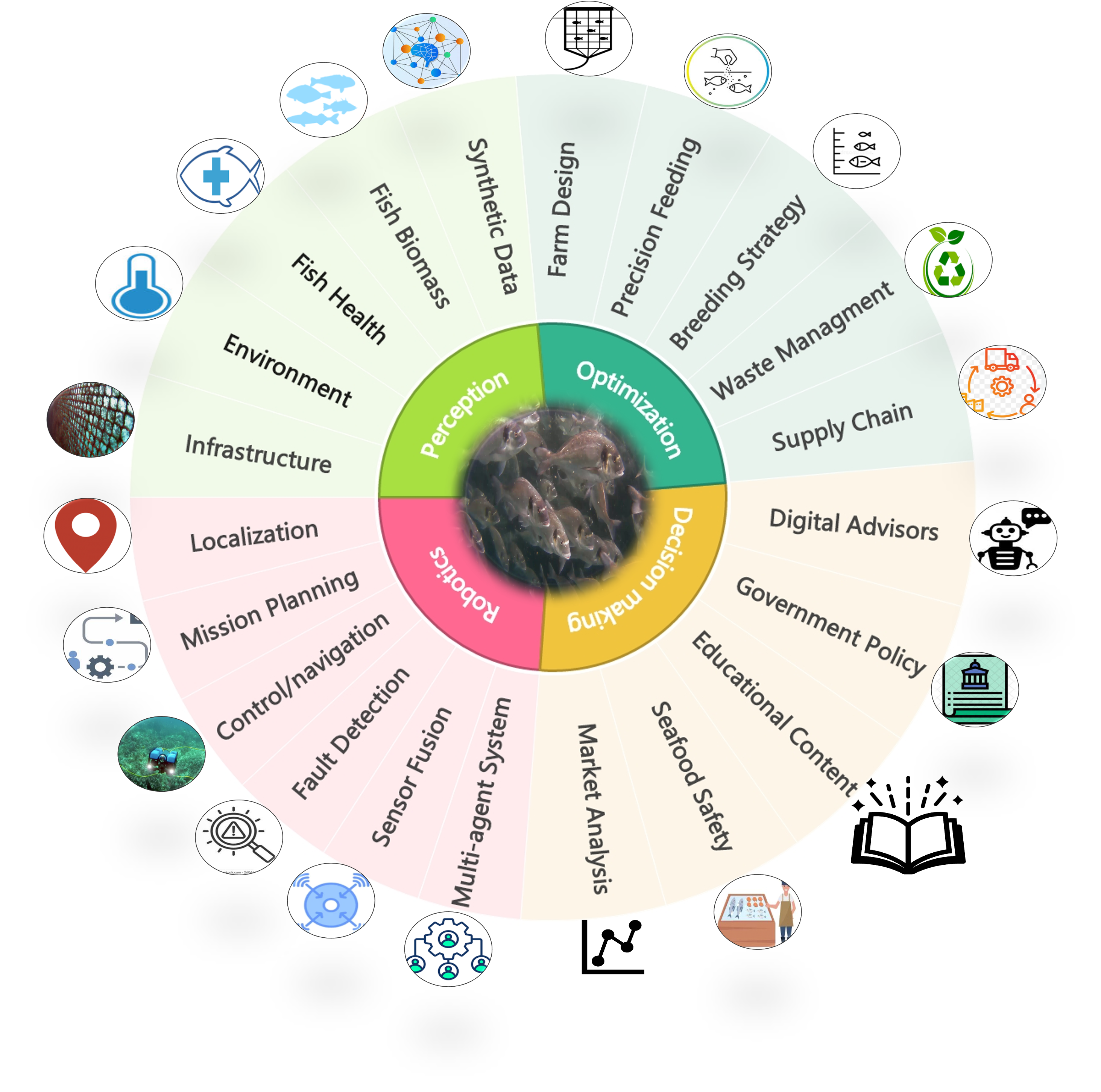}
    \caption{Applications of Generative Artificial Intelligence (GAI) in smart aquaculture, spanning automation, autonomous robotics, decision support, and optimization.}
    \label{fig:abs}
\end{figure}

To overcome these limitations, the industry must shift toward more adaptive and predictive frameworks capable of synthesizing complex data streams. Generative Artificial Intelligence (GAI) offers such transformative potential \citep{bommasani2021opportunities}. By enabling reasoning across modalities, including text, imagery, audio, and simulation, GAI facilitates novel use cases ranging from synthetic data generation for fish behavior modeling \citep{akkem2024comprehensive}, disease diagnosis, and feed optimization, to real-time decision-making via natural language interfaces \citep{vemprala2024chatgpt, aung2025artificial}. Foundation models such as ChatGPT, DALL·E, and diffusion-based generators are increasingly being applied as core building blocks for next-generation aquaculture platforms \citep{yang2023diffusion, wang2021intelligent}.

\begin{figure*}[t]
    \centering
    \includegraphics[width=1\linewidth]{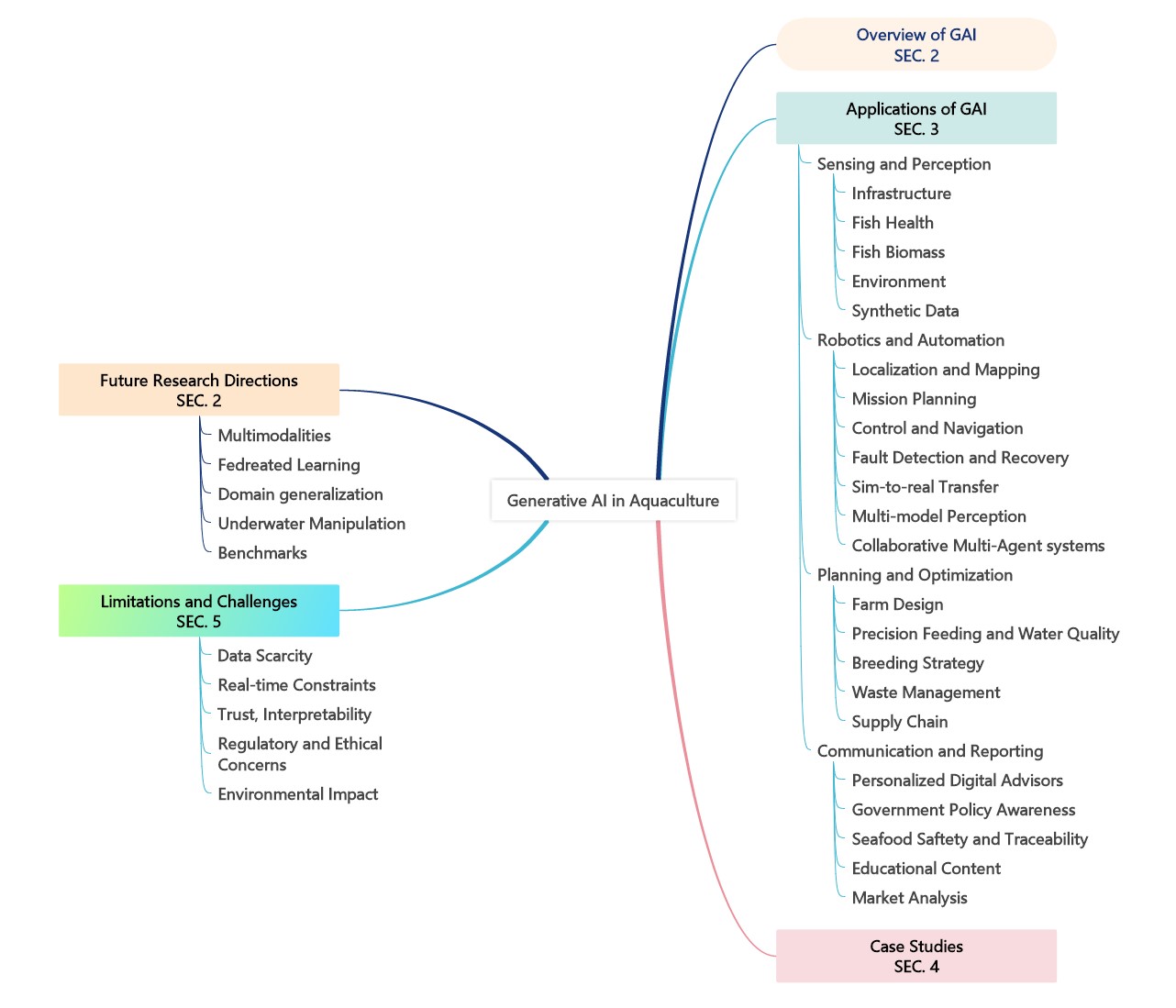}
    \caption{Structure of the review, organized around key thematic areas in which GAI contributes to aquaculture. The paper covers core models, applications, use cases, challenges, and future research directions.}
    \label{fig:rev-str}
\end{figure*}

As illustrated in Figure~\ref{fig:abs}, and further detailed in Table~\ref{tab:GAI-aquaculture}, GAI enables a wide array of capabilities across the aquaculture lifecycle including autonomous underwater inspection, multimodal sensor fusion, farm layout optimization, breeding strategy refinement, and blockchain-integrated traceability systems. GAI also plays an enabling role in robotic mission planning and fault recovery, supporting the transition toward highly autonomous and context-aware aquaculture operations \citep{ma2024survey, wen2025tinyvla}.

Despite increasing interest in generative technologies, their systematic integration in aquaculture remains limited. Prior reviews (e.g., \citep{fini2025application}) primarily focus on individual applications such as feed management or fish health, leaving a broader synthesis of robotics, planning, and digital transformation unexplored. To address this gap, this review offers the first comprehensive framework connecting GAI to all key layers of smart aquaculture from perception and automation to policy-level decision support.

\subsection*{Main contributions of this review}

\begin{itemize}
    \item We provide a technical overview of core GAI models, including diffusion models, vision-language foundation architectures, and retrieval-augmented generation (RAG) systems relevant to aquaculture engineering.
    
    \item We categorize and critically analyze diverse use cases where GAI addresses major aquaculture challenges, including infrastructure monitoring, water quality forecasting, breeding optimization, and stakeholder communication.
    
    \item We highlight real-world prototypes and emerging platforms, such as GAI-powered ROV mission planners, traceability systems integrated with blockchain, and multilingual advisory bots.
    
    \item We present a detailed tabular synthesis (Table~\ref{tab:GAI-aquaculture}) linking GAI models to their application domains, datasets, and evaluation benchmarks across aquaculture subsectors.
    
    \item We identify technical, ethical, and infrastructural barriers to deployment, and outline a roadmap for the scalable and responsible adoption of GAI in next-generation aquaculture systems aligned with Aquaculture 4.0.
\end{itemize}

The remainder of this paper is organized as follows, with its structure illustrated in Figure~\ref{fig:rev-str}. Section~\ref{sec:overview-GAI} provides a concise overview of GAI, while Section~\ref{sec:app-GAI} examines its applications within the aquaculture sector. Current research and case studies are discussed in Section~\ref{sec:cases}, followed by an exploration of existing limitations in Section~\ref{sec:limitations}. Future research directions are presented in Section~\ref{sec:future}, and the paper concludes with a summary of key findings in Section~\ref{sec:conclusion}.

\section{Overview of GAI} \label{sec:overview-GAI}

GAI has rapidly emerged as a transformative paradigm across scientific and engineering disciplines, shifting the role of AI from analytical reasoning to autonomous generation of coherent, multimodal content. These models learn and replicate the probabilistic structure of complex datasets to synthesize diverse outputs ranging from text and images to audio, video, simulation environments, and executable code~\citep{trigka2025evolution, balasubramaniam2024road}. 

Core GAI model architectures include diffusion models, transformer-based Large Language Models (LLMs), Generative Adversarial Networks (GANs), and Variational Autoencoders (VAEs). Each offers distinct strengths: GANs excel at generating high-fidelity imagery, VAEs provide interpretable latent spaces for structured content, diffusion models support progressive denoising for photorealistic outputs, and transformers deliver scalable, context-aware reasoning across modalities~\citep{li2025survey, aymen2024synthetic, openai2023chatgpt}.

\begin{figure}[tt]
    \centering
    \includegraphics[width=0.98\linewidth]{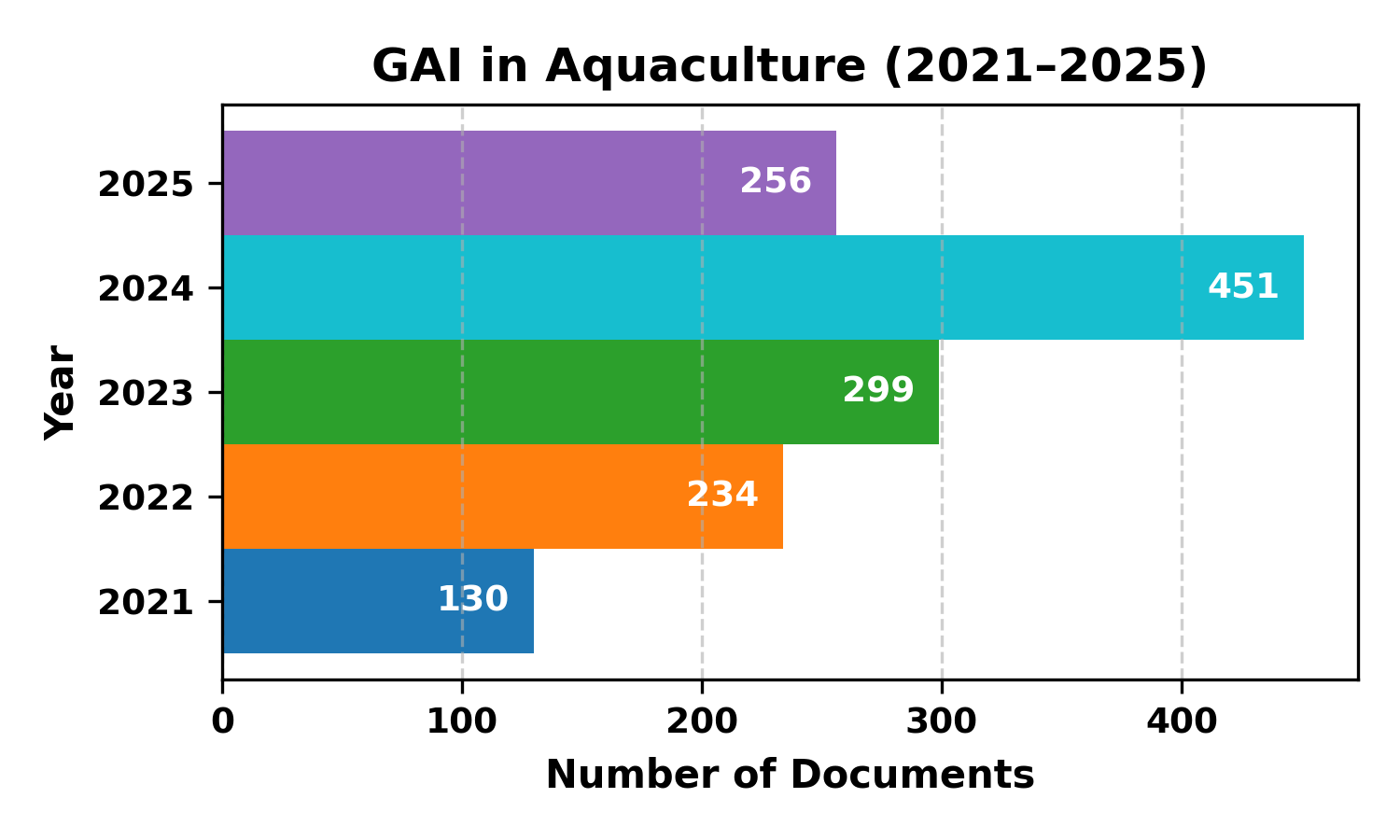}
    \caption{Growth of Scopus-indexed literature on GAI models and applications in the aquaculture sector from 2021 to 2025. Data retrieved in June 2025.}
    \label{fig:ai_aquaculture_growth}
\end{figure}

The increasing relevance of GAI to aquaculture is evidenced by a dramatic rise in related scientific literature over the past decade. As shown in Figure~\ref{fig:ai_aquaculture_growth}, the number of scopus-indexed publications containing the terms ``GAI'' or ``AI'' in Aquaculture, has grown from just 130 documents in 2021 to over 451 in 2024, with 256 already indexed in the first half of 2025. This upward trajectory reflects the growing convergence of intelligent automation, sustainability, and precision monitoring in the aquaculture domain~\citep{kim2024primer, rouzrokh2025medicine, croitoru2023diffusion}.

This section introduces the foundational GAI model families, outlines their underlying mechanisms, and contextualizes their potential in enabling intelligent, efficient, and ecologically responsible aquaculture systems.

\subsection{Architectures and uses of GAI}
\label{sec:core-gai}

GAI has emerged as a defining paradigm in modern AI, shifting the field from deterministic analytics toward creative, context-aware generation across diverse modalities. At its foundation, GAI involves models that learn the underlying probabilistic structure of input data and apply this understanding to autonomously synthesize outputs such as text, images, audio, video, and simulations. This shift has enabled a broad array of applications in fields ranging from robotics and healthcare to environmental science and education~\citep{trigka2025evolution, balasubramaniam2024road}.

GANs, VAEs, diffusion models, and LLMs are among the most foundational GAI architectures. GANs consist of two adversarial components, a generator and a discriminator, that engage in a dynamic training loop, enabling the creation of highly realistic outputs. These models have proven effective in tasks such as image-to-image translation, fish classification, and synthetic medical image generation~\citep{li2025survey, aymen2024synthetic}. VAEs, in contrast, adopt a probabilistic encoder-decoder approach, learning latent variable distributions to generate structured content. This property makes them particularly useful in environmental modeling, sonar signal synthesis, and structured acoustic data generation~\citep{pinheiro2021variational}.


Diffusion models are a more recent advancement in GAI. They generate high-resolution and semantically coherent content by progressively denoising a random noise distribution. Due to their ability to capture fine-grained structure, these models have been widely adopted in applications such as video generation, environmental simulation, and visual design~\citep{li2025design, aymen2024synthetic}. In parallel, transformer-based LLM, including GPT, Claude, and DeepSeek employ self-attention mechanisms that support long-range contextual reasoning and cross-modal generalization. These models are now used across diverse tasks, including text generation, code synthesis, multimodal dialogue, and simulation-based reasoning for medical and scientific domains~\citep{openai2023chatgpt, kim2024primer, mitra2025music}.

Cross-domain applicability is a hallmark of GAI. In healthcare, these models assist in diagnostic report generation, simulate rare medical scenarios, and automate clinical workflows~\citep{rouzrokh2025medicine, waisberg2025ophthalmology}. In robotics, GAI supports data augmentation for training autonomous agents, mission planning, and realistic environment synthesis for embodied learning~\citep{trigka2025evolution}. Environmental science also benefits, with models enabling street-level urban planning, crowd simulation, and water condition modeling for aquaculture and agriculture~\citep{huang2025walking}.

Despite these advances, several challenges remain. GAI systems are computationally intensive, often requiring significant resources for training and inference. They also raise concerns around interpretability, reproducibility, and potential misuse, especially when generating content in sensitive domains~\citep{kaswan2023review, balasubramaniam2024road}. Bias in training datasets can propagate into outputs, reinforcing harmful patterns unless actively mitigated. Furthermore, the generalization of these models across ecological, linguistic, or regulatory contexts requires additional research and domain-specific fine-tuning~\citep{kim2024primer}.

Nevertheless, GAI continues to evolve rapidly. Current directions include incorporating reinforcement learning with human feedback (RLHF), integrating multimodal context during generation, and developing energy-efficient architectures that democratize access to generative capabilities. As these models become more robust, controllable, and grounded in domain expertise, they are poised to become central tools in the digital transformation of sectors such as aquaculture, environmental monitoring, and precision medicine~\citep{trigka2025evolution, li2025design, kim2024primer}.

\subsection{Architectures in Practice}

GANs excel in producing synthetic images and augmenting datasets for fish classification and disease detection \citep{saxena2021generative}. VAEs support sonar signal reconstruction and structured scene synthesis \citep{pinheiro2021variational}. Diffusion models, such as Stable Diffusion \citep{rombach2022stablediffusion}, generate high-resolution underwater visuals through iterative denoising.

Transformer-based LLMs, including ChatGPT, Claude, and DeepSeek \citep{openai2023chatgpt, deepseek2024deepseekvl}, leverage attention mechanisms for coherent, interactive text generation. Models like BART and T5 further enable conditional and instruction-driven output generation. These are increasingly fine-tuned with RLHF, allowing for context-sensitive applications in aquaculture such as digital farm advisors or decision-support dashboards.

\subsection{Generative Modalities in Aquaculture }

Table~\ref{tab:gai_content_forms} categorizes the various forms of content that GAI systems can produce and maps them to their potential applications in aquaculture. This includes simulation of ROV trajectories, generation of advisory dialogues, synthesis of underwater inspection scenes, and multilingual alert systems. By aligning each content modality with relevant models and use cases, the table provides a comprehensive view of how GAI tools can be operationalized in real-world aquaculture scenarios.

\begin{figure*}[t]
    \centering
    \includegraphics[width=1\linewidth]{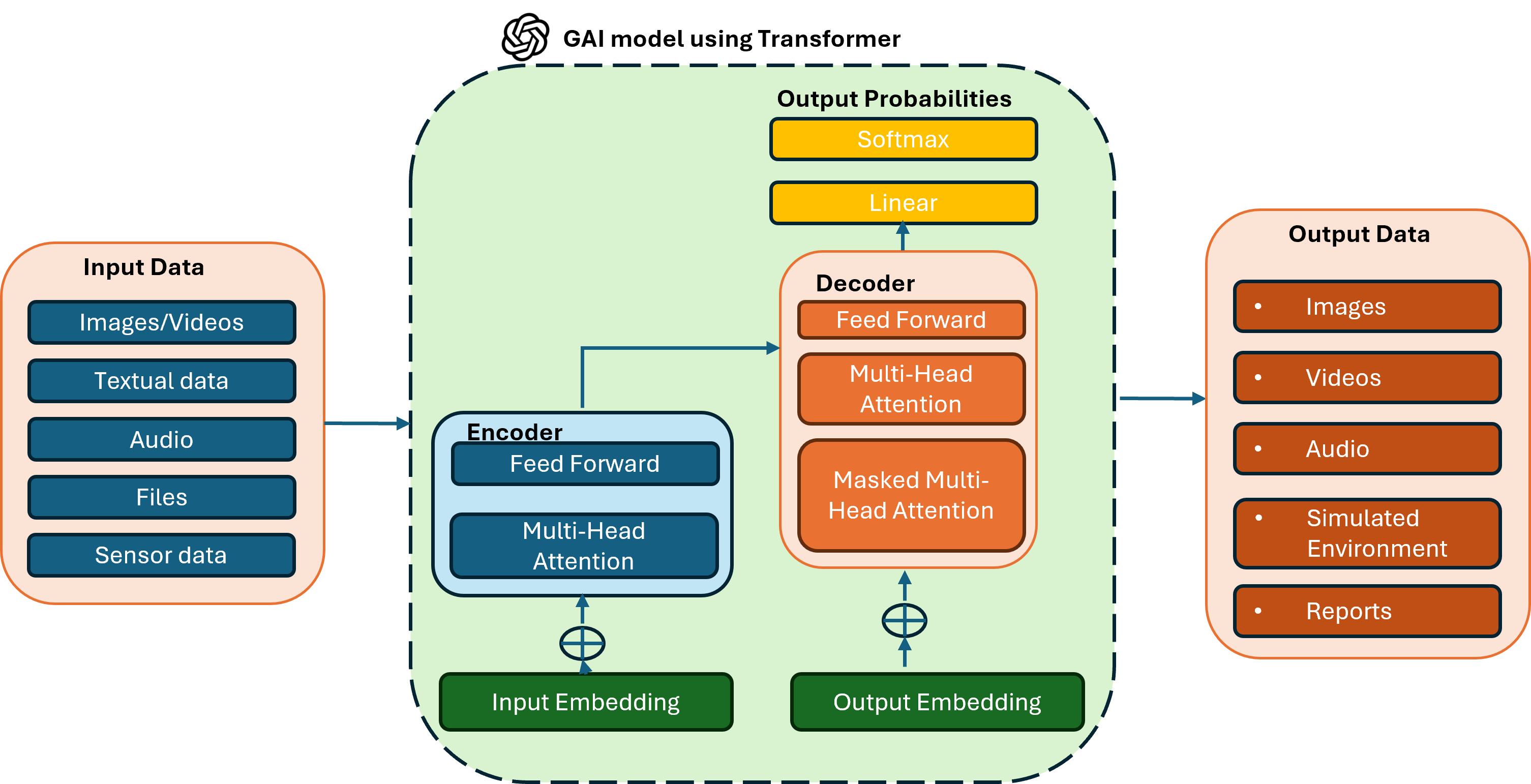}
    \caption{Generalized architecture of a Generative AI system. Diverse input modalities such as images, text, audio, and sensor data are processed through an encoder-decoder transformer. Outputs include images, simulations, audio, and text reports.}
    \label{fig:rev-gai}
\end{figure*}

\begin{table*}[t]
\centering
\caption{Overview of generative outputs and associated aquaculture applications enabled by GAI models.}
\begin{tabular}{p{4cm}p{8cm}p{3.5cm}}
\toprule
\textbf{Type of Content} & \textbf{Aquaculture Application} & \textbf{GAI Models / Architectures (Refs)} \\
\midrule
Text / chat & Interactive digital advisors, inspection summaries, compliance documents & ChatGPT, Claude, DeepSeek \citep{openai2023chatgpt, deepseek2024deepseekvl} \\
Image Generation & Net damage visualization, fish disease rendering, environmental design & DALL·E, Stable Diffusion, Midjourney \citep{ramesh2021dalle, rombach2022stablediffusion} \\
Video Generation & Simulation of fish schooling, ROV path planning, farm visualization & Gen-2 (Runway), Veo (DeepMind), Sora (OpenAI) \citep{croitoru2023diffusion} \\
Audio / Sound & Underwater sound simulation, alert generation, acoustic behavior cues & Bark, AudioGen, Voicebox \citep{croitoru2023diffusion} \\
Simulation / Environment & Virtual net testing, environmental change modeling, system control loops & GAIA-1, FlowDreamer, World Models \citep{croitoru2023diffusion} \\
Code / Automation Logic & ROS node generation for perception and control, automation scripts & Codex, RoboCoder, CodeGemma \citep{li2024foundation} \\
Document / PDF Analysis & Summarization of policy/regulatory documents, report generation & GPT-4 with vision, Claude-Opus, Donut \citep{vemprala2024chatgpt} \\
Image Analysis & Underwater infrastructure inspection, anomaly detection, fish tracking & VAE, GANs, Mamba \citep{saxena2021generative, pinheiro2021variational, ma2024survey} \\
\bottomrule
\end{tabular}
\label{tab:gai_content_forms}
\end{table*}

\subsection{Opportunities and Challenges in Aquaculture}

In aquaculture, GAI enables high-impact solutions across several fronts: bridging data scarcity through synthetic augmentation, simulating rare ecological events, automating reporting, and enhancing multimodal monitoring \citep{ma2024survey, li2024foundation}. For instance, GAI models can autonomously generate inspection reports from underwater footage, or simulate disease progression scenarios under varying water parameters. These tools also facilitate the design of personalized advisory systems and scalable control mechanisms for robotic aquaculture.

However, several challenges remain, such as ensuring generalization across diverse aquatic environments, achieving real-time inference under limited bandwidth, and validating synthetic outputs ecologically. These issues are further discussed in Section~\ref{sec:limitations}.

\section{Applications of GAI in Aquaculture}\label{sec:app-GAI}

GAI is rapidly transforming aquaculture by enabling intelligent, multimodal, and data-driven solutions across the entire value chain. Its applications span from underwater robotics and precision farming to market analytics and education, bridging key functional areas such as sensing, decision support, planning, automation, and communication. 

As summarized in Table~\ref{tab:GAI-aquaculture}, recent advancements demonstrate GAI's capability to enhance infrastructure inspection using drone-based vision models \citep{ubina2021drones}, support predictive fish health diagnostics with deep learning and AIoT frameworks \citep{roy2025ai, yang2025aidriven}, and improve biomass estimation and environmental monitoring through deep learning and LLMs \citep{kong2025aasnet, sundaravadivel2024llms}. Beyond perception, GAI also empowers robotics applications in mission planning \citep{li2024foundation}, digital twin simulation \citep{fini2025generative}, and multi-agent coordination \citep{fortino2025generative}. In the realm of optimization, GAI-driven frameworks have been applied to feeding and breeding strategy modeling \citep{aung2025review}, and waste treatment scenario simulation \citep{ragab2025overview}. Furthermore, generative models and LLMs are being integrated into advisory tools \citep{jasmin2024chatbot}, educational platforms \citep{bhusan2025it}, and blockchain-based traceability systems \citep{zhang2021biotts, narang2024mussels}, underscoring GAI’s role in communication, transparency, and regulatory compliance. 

Despite their diverse roles, these applications remain largely in the prototyping or early deployment stage, emphasizing the need for continued interdisciplinary development and context-aware validation. Table~\ref{tab:GAI-aquaculture} offers a consolidated view of representative use cases, underlying models, datasets, and evaluation protocols that collectively define the current frontier of GAI-enabled aquaculture.

\begin{table*}[ht]
\caption{Summary of recent research on GAI applications in aquaculture sensing and perception.}
\centering
\begin{tabular}{p{4.2cm} p{3.6cm} p{3.8cm} p{4.2cm}}
\toprule
\textbf{Models (Refs)} & \textbf{Use Case} & \textbf{Datasets} & \textbf{Evaluation Methods} \\
\midrule
Deep learning, drone-based visual models \citep{ubina2021drones} & Infrastructure Inspection & Aerial image datasets & Detection accuracy, visual inspection reliability \\
CNNs, LSTMs, predictive and AIoT frameworks \citep{roy2025ai, roy2024smart, yang2025aidriven, krivoguz2025epizootic, shang2025health} & Fish Health Diagnostics & Real-time sensors, historical logs & Real-time performance, predictive accuracy \\
AASNet, YOLOv8 fish detection \citep{kong2025aasnet, zhang2025uav, zhang2024counting} & Biomass and Growth Estimation & UIIS, USIS, hybrid image sets \citep{steele2024virtual} & mAP score, growth accuracy, inference speed \\
Multimodal LLMs, sparse attention \citep{sundaravadivel2024llms, arepalli2024water, kanwal2024iot} & Environmental Monitoring & Water quality logs, sensor inputs & Risk prediction, classification accuracy \\
Generative AI, hybrid augmentation \citep{bendel2024whisperer, fini2025generative} & Synthetic Data Generation & Synthetic coral/fish datasets \citep{steele2024virtual} & FID, generalization, augmentation impact \\
AIoT-enhanced deep models for SLAM \citep{huang2025aiot} & Localization and Mapping & IoT sensor data, underwater imaging & Localization accuracy, SLAM stability, mapping robustness \\
Foundation models for resource optimization \citep{li2024foundation} & Mission Planning & Smart aquaculture systems, operational logs & Route planning efficiency, scheduling optimization \\
Interactive generative AI agents \citep{fortino2025generative, fortino2024generative} & Control and Navigation & LLM-driven robots, human-robot interaction logs & Command accuracy, latency, operator usability \\
ML/DL models for predictive diagnostics \citep{roy2025ai, huang2025aiot} & Fault Detection and Recovery & Real-time fault/event data & Fault classification accuracy, recovery time \\
Digital twin simulators with generative AI \citep{huang2025aiot, fini2025generative} & Digital Twin & Virtual replicas, real-time farm simulations & Predictive reliability, system response simulation \\
Multimodal foundation models for fusion \citep{huang2025aiot, li2024foundation} & Multimodal Perception and Sensor Fusion & Sensor data (pH, DO, temp, vision) & Cross-modal fusion accuracy, sensing delay, robustness \\
Collaborative AI agents \citep{fortino2025generative, fortino2024generative} & Collaborative Multi-Agent Systems & Coordinated robotic systems & Task coordination success rate, distributed efficiency \\
Generative AI, sensor-integrated layout planning \citep{fini2025generative, srikanth2024ai} & Farm Design and Infrastructure Planning & Real-time sensor data, integrated resource systems & Land use optimization, resource efficiency, layout accuracy \\
AI/ML-based behavioral models, DL water quality predictors \citep{roy2025ai, ragab2025overview, srikanth2024ai} & Precision Feeding and Water Quality Simulation & Feeding logs, water quality sensors, camera data & Feed conversion ratio, WQ prediction accuracy, nutrient waste \\
Predictive analytics on genetics, generative augmentation \citep{yang2025aidriven, aung2025review, rafiq2025generative} & Breeding Strategy Optimization & Genetic data, breeding history, synthetic trait sets & Productivity, health index, genetic diversity scores \\
AI-based waste prediction and scenario simulation \citep{ragab2025overview, fini2025generative} & Waste Management Optimization & Waste output logs, simulation models & Waste volume reduction, treatment efficiency, ecological impact \\
Generative AI for demand forecasting, logistics optimization \citep{pallottino2025applications, aung2025review} & Aquaculture Supply Chain Optimization & Market data, logistics records, synthetic consumer trends & Delivery time, cost reduction, market match accuracy \\
LLMs, AquaGent chatbot \citep{jasmin2024chatbot} & Personalized Digital Advisors & Facebook Messenger, Telegram bots & Response accuracy, user satisfaction \\
LLMs + Blockchain + IoT \citep{zhang2021biotts, gogou2020blockchain, narang2024mussels} & Traceability and Transparency & BIOT-TS, Mussels IoT Tracker & Data integrity, compliance verification, traceability rate \\
LLMs for educational content and AI-enhanced collaboration \citep{bhusan2025it, sevin2025ecosystem} & Aquaculture Training and Research & Smart learning tools, virtual ChatGPT environments & Learner engagement, training efficiency \\
Generative AI + ChatGPT \citep{weichelt2024chatgpt, fini2025generative} & Market Analysis and Reporting & Text generation, market reports, predictive tools & Forecast accuracy, analysis time saved \\
LLMs for disease classification, reporting, data extraction \citep{qiao2025tdaglm, nugraha2024extraction, li2025textclass, jiang2023map} & Automated Aquaculture Reporting & TDA-GLM, TABEM, knowledge graphs & Text classification F1-score, extraction accuracy \\
LLMs for climate-aware decision support \citep{leite2025rag} & Environmental Intelligence & Retrieval-Augmented Generation (RAG) & Report precision, response time \\
LLMs for conversational analysis \citep{pugh2023conversations} & Collaborative Intelligence & Dialog modeling, sustainability discussions & Dialog act precision, innovation impact \\

\bottomrule
\end{tabular}

\label{tab:GAI-aquaculture}
\end{table*}

\subsection{Sensing and Perception}

\begin{figure}[t]
\centering
\subfloat[ROV.  \label{fig:B2}]{\includegraphics[width=0.5\columnwidth, height=0.3\columnwidth]{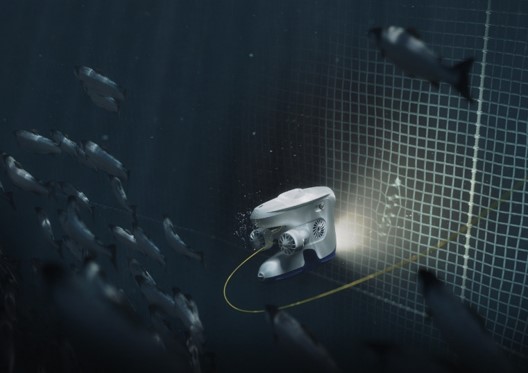}}
\subfloat[Net holes. \label{fig:B1}]{\includegraphics[width=0.5\columnwidth,height=0.3\columnwidth]{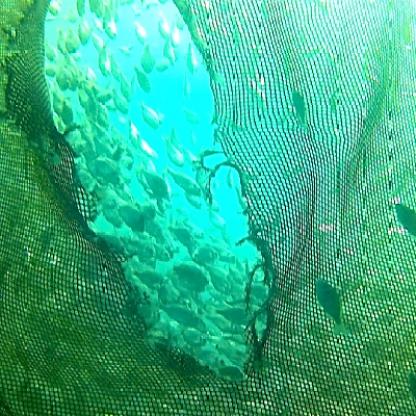}}
\vspace{5pt}\hspace{0.2cm}
\subfloat[Biofouling. \label{fig:B3}]{\includegraphics[width=0.5\columnwidth]{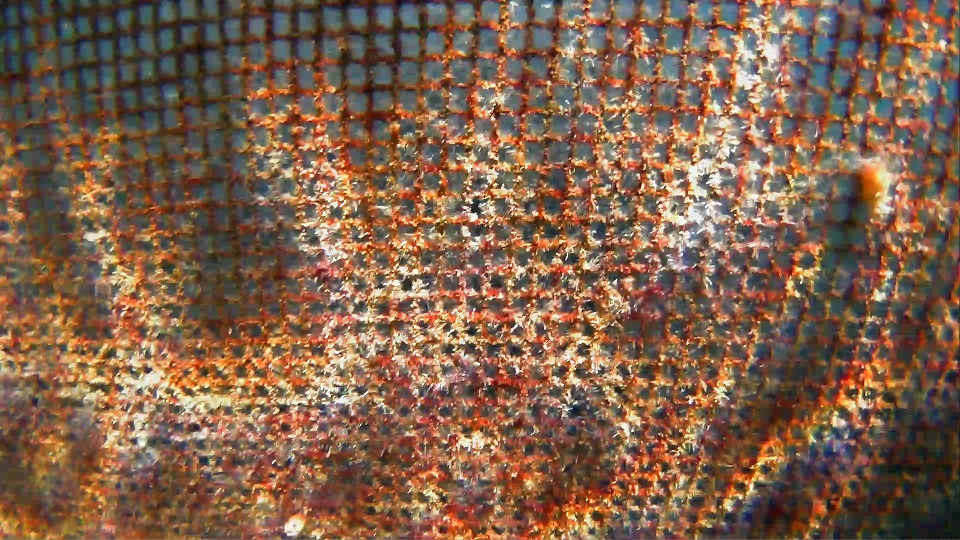}}
\subfloat[Vegetation.  \label{fig:B4}]{\includegraphics[width=0.5\columnwidth]{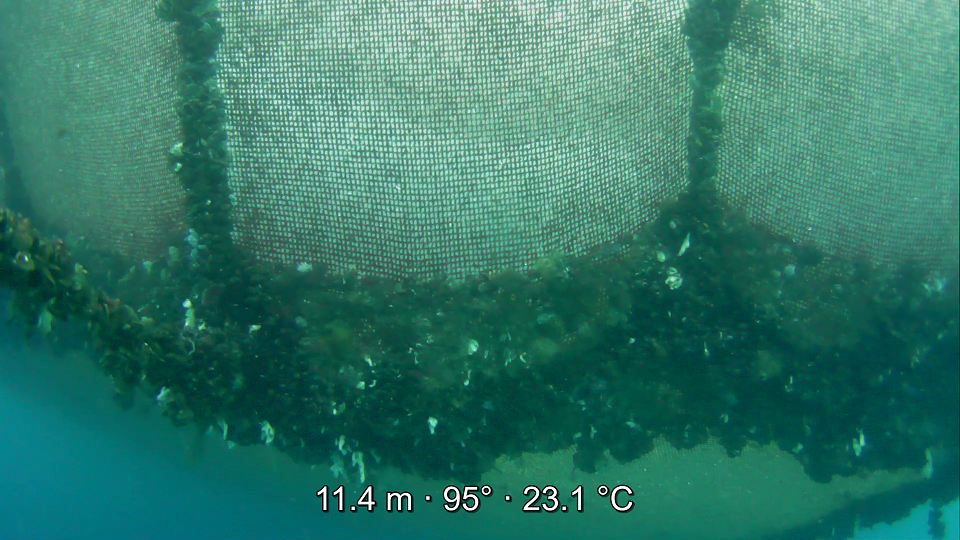}}\vspace{5pt}\hspace{0.2cm}
\caption{Automatic aquaculture inspection using ROV. (a) shows an ROV during aquaculture net pens inspection, (b) shows an example of the hole on the net, (c) shows an example of biofouling defects on the net, and (d) shows an example of vegetation attached to the net. }
\label{fig:biofouling}
\end{figure}

As aquaculture continues to evolve into a data-intensive industry, sophisticated sensing and perception systems have become essential to enhance operational efficiency, sustain fish health, and achieve environmental sustainability \citep{subasinghe2009global}. GAI and foundation models have emerged as transformative technologies within this domain, significantly advancing data interpretation, predictive analysis, and scenario synthesis. Recent developments highlight the substantial contributions of GAI to aquaculture sensing and perception, particularly in infrastructure monitoring, fish health diagnostics, biomass estimation, environmental monitoring, and synthetic data augmentation. For instance, Figure \ref{fig:biofouling} illustrates the use of an ROV for routine inspection of aquaculture net pens to detect net damage, biofouling, and vegetation growth. This example highlights the sensing and perception capabilities that can be substantially augmented by GAI, particularly in improving the interpretation and understanding of complex imaging data \citep{rahman2021integrated}.

\subsubsection{Infrastructure Inspection and Monitoring}

Regular and precise monitoring of aquaculture infrastructure is critical to maintaining structural integrity and ensuring optimal operational efficiency. However, conventional manual inspection methods are often labor-intensive, costly, and susceptible to human error, especially in harsh underwater conditions characterized by poor visibility and environmental disturbances. GAI addresses these challenges by employing deep generative models, including GANs, ChatGPt and diffusion models, to create realistic synthetic scenarios for automated inspection training \citep{wang2021intelligent}.

\begin{figure}[t]
    \centering
    \includegraphics[width=1\linewidth]{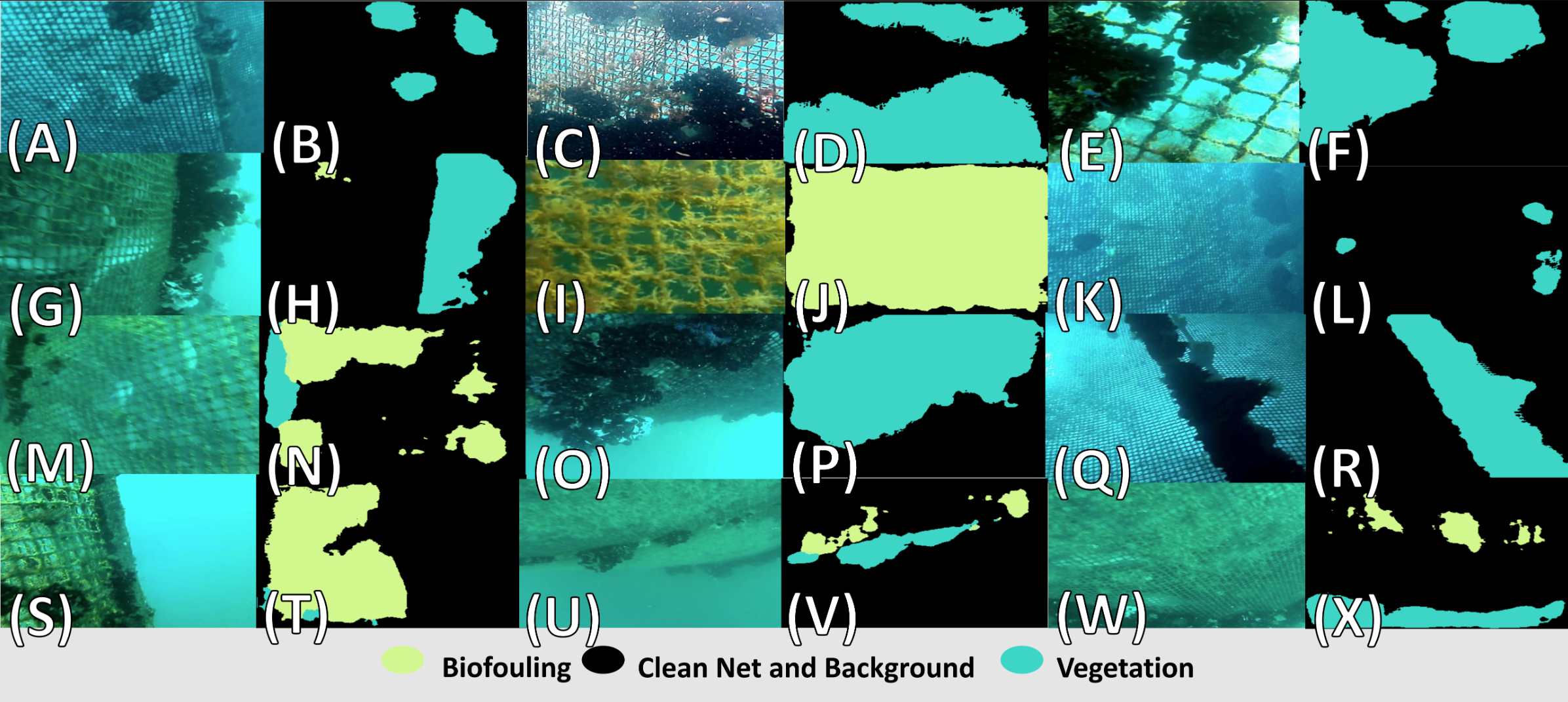}
    \caption{ Automated segmentation of underwater aquaculture net images, showing raw images and corresponding masks for biofouling (light green), clean net and background (black), and vegetation (cyan) to support accurate monitoring and data interpretation. Image courtesy of \citep{akram2024aquaculture}.}
    \label{fig:ne}
\end{figure}

Recent studies demonstrate how GAI models synthesize realistic underwater images depicting various infrastructure conditions, such as net pen degradation, biofouling accumulation, and structural damage. For example, generative diffusion models have been used successfully to simulate diverse biofouling conditions on fish cages, enabling inspection robots trained on these synthetic datasets to robustly detect real-world anomalies with high precision \citep{chai2023deep, panetta2021comprehensive}. Additionally, foundation models facilitate the integration of diverse sensor data such as acoustic and visual inputs enhancing automated detection capabilities and reducing dependency on costly human led inspections \citep{gonzalez2023integration, elmezain2025advancing}. As shown in Figure~\ref{fig:ne}, automated segmentation enables clear differentiation between biofouling, clean net areas, and vegetation for improved net pen monitoring.

\subsubsection{Fish Health Diagnostics and Monitoring}

Effective fish health monitoring and early disease detection are vital for preventing outbreaks and minimizing economic losses in aquaculture operations \citep{assefa2018maintenance}. GAI significantly enhances diagnostic capabilities by generating synthetic scenarios of fish diseases and anomalous behavioral patterns. These artificially generated scenarios improve the training of diagnostic models, leading to robust real-time monitoring solutions.

In recent applications, deep learning been effectively utilized to synthesize realistic visual data representing diseases such as sea lice infestations, fungal infections, or bacterial diseases \citep{bondad2005disease}. By training automated diagnostic models on these synthetically enriched datasets, the accuracy and reliability of automated visual diagnostic systems have notably improved \citep{kim2018autoencoder, gupta2022accurate, li2022advanced}. Moreover, GAI-driven models facilitate continuous health assessments by accurately predicting disease progression, thus enabling proactive interventions and significantly improving animal welfare and farm productivity.

\subsubsection{Fish Biomass and Growth Estimation}

Accurate fish biomass estimation and growth rate monitoring are fundamental to precision aquaculture, directly influencing feeding strategies, harvesting schedules, and economic forecasting \citep{zhang2024fully}. Traditional methods typically involve invasive sampling techniques or manual estimations prone to inaccuracies. GAI transforms this process by synthesizing realistic fish imagery and growth-stage scenarios under diverse environmental conditions, substantially improving biomass estimation accuracy and robustness.

\begin{figure}[t]
    \centering
    \includegraphics[width=1\linewidth]{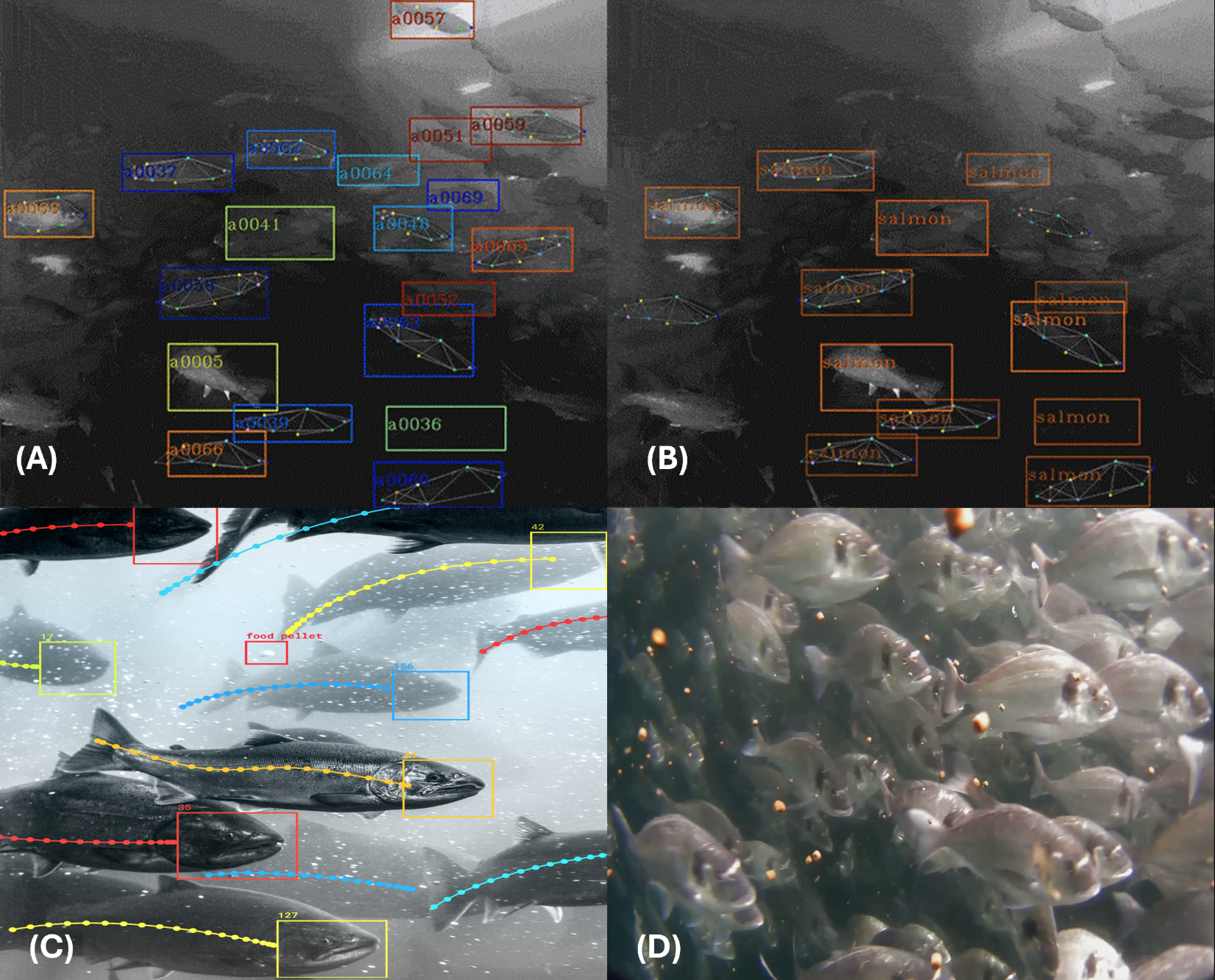}
    \caption{ Examples of fish detection, tracking, and biomass estimation in aquaculture: (A) multi-species detection with bounding boxes, (B) single-species detection, (C) fish tracking with trajectory paths, and (D) biomass estimation in dense fish schools. Images courtesy of \citep{ieee_aquaculture_2024,innovasea_biomass_2025,aquaculture_north_america_google_2020}}
    \label{fig:fishbiomass}
\end{figure}

Deep learning models have been leveraged to simulate fish populations at various growth stages, environmental conditions, and density levels, enabling comprehensive training of predictive biomass estimation models. These techniques help overcome challenges posed by variable underwater visibility, sensor noise, and inconsistent data annotations, ensuring greater reliability and generalization of biomass monitoring systems \citep{zhang2024fully, abinaya2022deep}. By augmenting real-world datasets with extensive synthetic scenarios, GAI greatly enhances the predictive accuracy of biomass models, ultimately facilitating better operational planning and resource management \citep{zahir2024review}. An example illustrating fish detection, tracking, and biomass estimation in typical aquaculture settings is shown in Figure~\ref{fig:fishbiomass}.

\subsubsection{Environmental Condition Monitoring}

Effective environmental monitoring is crucial for maintaining optimal growth conditions and mitigating environmental impacts within aquaculture systems \citep{english2024review}. GAI techniques play a significant role by synthesizing diverse environmental scenarios, including variations in critical parameters such as dissolved oxygen levels, pH fluctuations, turbidity, and nutrient concentrations. These simulated scenarios enable predictive analyses that support proactive management decisions by aquaculture operators.

\begin{figure}[t]
    \centering
    \includegraphics[width=1\linewidth]{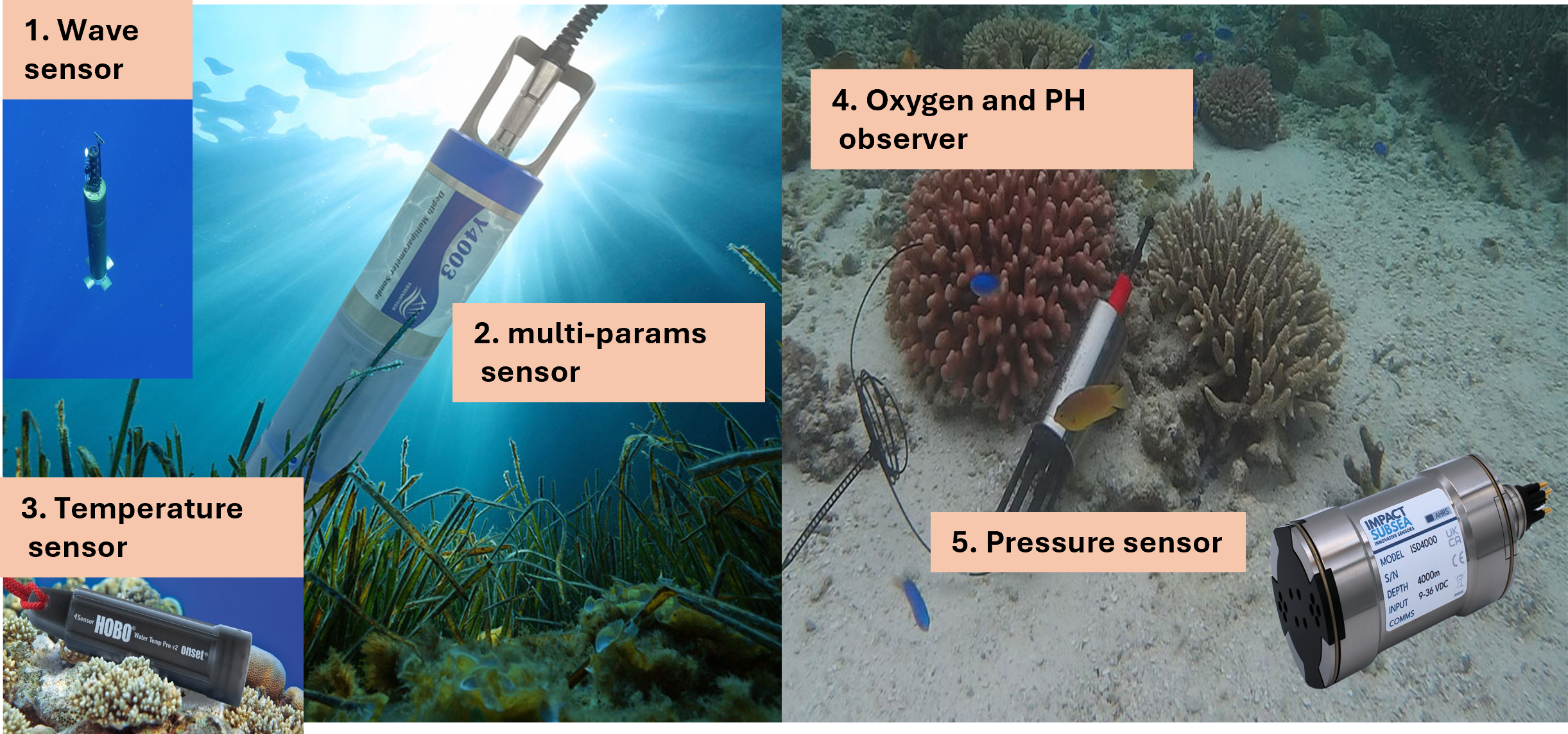}
    \caption{Common environmental sensors used in aquaculture monitoring: (1) wave sensor, (2) multi-parameter sensor, (3) temperature sensor, (4) oxygen and pH observer, and (5) pressure sensor. Images courtesy of \citep{yosemitech_multi_sensor,tempcon_temperature_sensor,ecomagazine_ph_oxygen_sensor,oceanscience_pressure_sensor}.}
    \label{fig:envmon}
\end{figure}

For instance, generative models have successfully simulated environmental conditions leading to harmful algal blooms or hypoxic events, allowing predictive models to forecast the occurrence of these critical environmental conditions accurately \citep{ren2020research, eze2020dissolved}. As a result, aquaculture managers can implement preventive actions, such as targeted aeration adjustments or controlled feeding strategies, before negative conditions manifest. Such predictive environmental management substantially improves overall operational resilience, sustainability, and regulatory compliance \citep{pachaiyappan2024enhancing}. Examples of environmental monitoring sensor modules are illustrated in Figure~\ref{fig:envmon}.

\subsubsection{Synthetic Data Generation and Data Augmentation}

A significant obstacle in aquaculture perception tasks is the scarcity of high-quality, diverse, and accurately annotated datasets, particularly in challenging underwater environments characterized by dynamic conditions, poor visibility, and limited accessibility \citep{li2024data}. GAI provides a highly effective solution by generating extensive synthetic datasets, significantly enriching training data and facilitating effective model training across diverse perception tasks \citep{li2024foundation,ls1}.

Recent research emphasizes the efficacy of GANs and diffusion models in creating realistic synthetic images and sensor data for tasks such as object detection, semantic segmentation, and anomaly detection in aquaculture systems. Synthetic data generation using generative models addresses data diversity limitations by realistically simulating varying underwater visibility conditions, infrastructure anomalies, fish behavior, and health scenarios \citep{natarajan2024synth,ls2}. Consequently, perception models trained on these extensive synthetic datasets exhibit improved generalization capabilities, significantly reducing reliance on labor intensive manual data collection and annotation efforts. This approach leads to more efficient, robust, and reliable perception systems, ultimately enhancing operational effectiveness and reducing operational risks in aquaculture settings \citep{ls3}.

\subsection{Robotics and Automation}

\begin{figure}[t]
    \centering
    \includegraphics[width=1\linewidth]{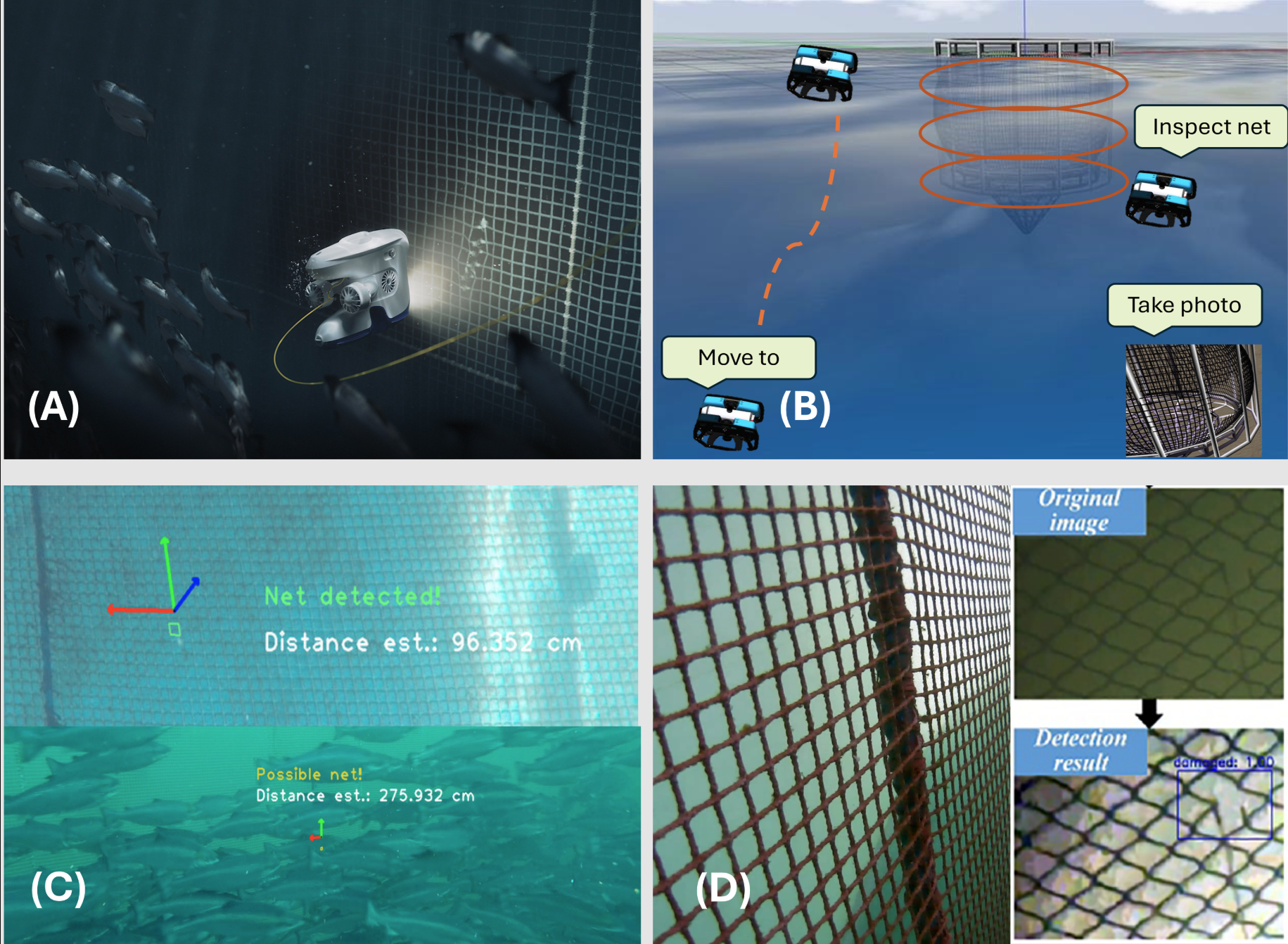}
    \caption{Examples of robotics and automation applications in aquaculture: (A) underwater robot inspecting a fish net, (B) autonomous navigation and inspection planning, (C) real-time net detection and distance estimation, and (D) net defect detection and annotation.}
    \label{fig:robonet}
\end{figure}

As aquaculture operations scale in complexity and volume, robotics and automation have emerged as key enablers for improving efficiency, safety, and sustainability. Traditional manually operated systems are increasingly being replaced with autonomous platforms equipped with AI-driven perception, planning, and control capabilities. These robotic systems are deployed to carry out critical tasks such as net pen inspection, biofouling removal, fish monitoring, and environmental sensing in conditions that are hazardous or impractical for human divers. The dynamic and unstructured nature of underwater environments, including poor visibility, variable currents, and biofouling, necessitates robust robotic platforms capable of adaptive and intelligent behavior \citep{ra3}. An illustrative example of robotics applications in aquaculture is shown in Figure~\ref{fig:robonet}, highlighting underwater inspection, autonomous navigation, and net defect detection.

A diverse array of robotic platforms is currently deployed across aquaculture operations, with ROVs playing a central role in underwater inspection and maintenance. Commonly used ROVs such as the BlueROV2, Argus Mini, and Blueye ROV are outfitted with high-resolution cameras, robotic arms, and environmental sensors to support tasks like net pen inspection, biofouling assessment, and sample collection. These vehicles are typically tethered and controlled by human operators, though many are increasingly integrated with onboard autonomy for semi-automated mission execution. Beyond ROVs, surface-level operations are supported by Unmanned Surface Vehicles (USVs) like the SEABOTS SB100, H2Omni-X \citep{H2OmniX2021}, and AutoNaut, which autonomously perform water quality monitoring and surveillance across large farm areas \citep{ra2}.

Despite these advances, most current robotic systems operate based on pre-defined scripts or reactive controllers with limited capacity for real-time decision-making. The integration of GAI introduces a transformative shift, enabling robots to not only sense and act, but also to reason, plan, and communicate \citep{wen2025tinyvla}. By embedding generative models within robotic architectures, these platforms can achieve higher levels of autonomy, learning from prior missions, simulating future outcomes, and generating context-aware behaviors tailored to specific aquaculture needs \citep{ra1,heshmat2025underwater}.

\subsubsection{Localization and Mapping}

Localization and mapping represent persistent challenges within aquaculture, particularly due to complex underwater environments characterized by dynamic conditions, low visibility, and limited sensor reliability \citep{boer2023deep}. GAI addresses these challenges by significantly enhancing Simultaneous Localization and Mapping (SLAM) techniques through realistic scenario generation and predictive modeling \citep{merveille2024advancements}. Generative models can synthesize detailed underwater maps and environments from limited sensor data, enabling robotic platforms to navigate reliably despite partial or noisy sensory inputs \citep{chakravarty2019gen}.

For example, diffusion-based generative models trained on underwater imaging datasets have successfully reconstructed high-resolution maps from sparse and degraded sonar data, substantially improving underwater navigation accuracy \citep{safron2022generalized,liu2024precise}. Such synthesized environments also facilitate extensive testing and validation of localization algorithms, leading to more robust and adaptive robotic operations. These advancements are critical in environments such as offshore net pen farms and submerged infrastructure, where accurate positioning is essential for efficient operational planning and management \citep{rahimi2025generative}.

\subsubsection{Mission Planning}

Effective mission planning for ROVs in aquaculture monitoring requires continuous adaptation to dynamic underwater conditions, including turbidity, marine growth, structural changes, and battery constraints. GAI, particularly large foundation models, introduces a paradigm shift by enabling predictive, adaptive, and context-aware planning. These models can generate hypothetical mission scenarios, simulate environmental disturbances, and propose optimal action sequences, allowing ROVs to adjust inspection strategies in real time \citep{andreoni2024enhancing, swinton2025autonomous}.

Recent advances, such as MissionGPT, uses transformer-based architectures to generate symbolic mission plans from textual prompts, which can be used to schedule multi-stage inspections, prioritize high-risk areas, or replan tasks based on unexpected environmental inputs \citep{berman2024missiongpt}. Similarly, Diffusion-based planners can simulate degraded visibility conditions or sensor failures to train ROVs on robust policy selection under uncertainty. In aquaculture settings, these capabilities allow proactive scheduling of net inspections by forecasting biofouling progression, adjusting inspection frequencies, or re-routing paths to avoid thruster overload near strong currents \citep{siderska2024complementing}.

Moreover, models like \citep{yang2023oceanchat} demonstrate how LLMs can translate high-level user intents (e.g., ``Inspect the southwest net corner if visibility is above 40\%'') into low-level action sequences grounded in the current environment state allowing natural language-driven, adaptive control for ROV operations. These innovations enhance inspection reliability, reduce manual intervention, and ensure energy-efficient navigation tailored to real-time aquaculture demands.

\subsubsection{Control and Navigation}

GAI plays a pivotal role in robotic control and navigation within aquaculture environments, which are frequently impacted by unpredictable disturbances such as water currents, tidal fluctuations, and obstacles. Foundation models integrated with reinforcement learning algorithms allow robots to autonomously develop sophisticated navigation strategies capable of responding to these dynamic disturbances \citep{isreal2025rise}.

GAI supports the synthesis of realistic simulation environments, allowing robotic systems to experience numerous variations of real-world challenges during training phases. For example, reinforcement learning models trained with generative adversarial data augmentation have shown significant improvements in navigation performance, enabling autonomous vehicles to rapidly adapt and generalize learned policies to unseen, real-world scenarios. This ensures robust operational performance in challenging conditions commonly encountered in aquaculture environments, such as offshore cage inspections and precise feeding operations \citep{firoozi2025foundation,serpiva2025racevla}.

\subsubsection{Fault Detection and Recovery}

Robust fault detection and rapid recovery mechanisms are crucial for maintaining continuous operational reliability in aquaculture robotics \citep{nagothu2025advancing}. GAI significantly enhances these capabilities by proactively identifying potential faults through predictive modeling and scenario simulation \citep{mikolajewska2025generative}. Foundation models trained on comprehensive historical operational data can simulate diverse fault scenarios, sensor degradations, and mechanical failures, thus preparing robotic systems for swift autonomous detection and diagnostic responses \citep{hu2025leveraging}.

Through generative predictive maintenance models, aquaculture robots, such as automated feeding systems and ROVs, can continuously monitor sensor health and operational parameters, preemptively identifying anomalies indicative of impending equipment failures \citep{mugala2025leveraging}. This proactive approach substantially reduces downtime by enabling immediate corrective actions, maintaining system reliability, and ensuring seamless aquaculture operations \citep{mahale2025comprehensive,huang2025artificial}.

\subsubsection{Digital Twin}
\begin{figure}[t]
\centering
\includegraphics[width=1\linewidth]{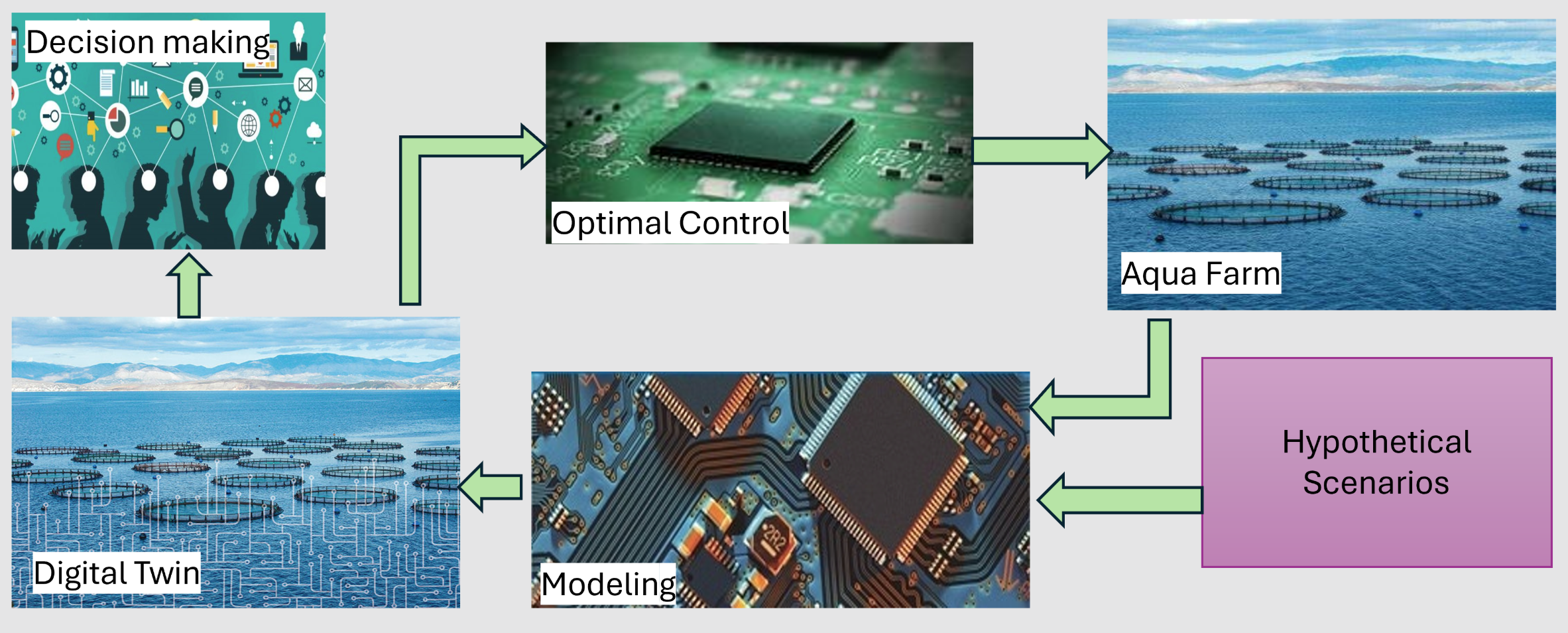}
\caption{Illustration of a Digital Twin framework enhanced with GAI for aquaculture robotics. The virtual model simulates real-time conditions and hypothetical scenarios to support optimal control, decision-making, and predictive analytics.}
\label{fig:digittwin}
\end{figure}

A critical challenge in aquaculture robotics is ensuring that virtual simulations and real-world deployments are tightly synchronized \citep{kargar2024emerging}. GAI plays a pivotal role in enhancing digital twins (DT) by generating high-fidelity synthetic data, simulating unobserved environmental conditions, and modeling complex interactions in aquaculture setups. By integrating GAI with DT systems, operators and autonomous agents can conduct scenario based training, fault injection analysis, and predictive planning using a continuously updated virtual replica of the aquafarm as shown in Figure~\ref{fig:digittwin} \citep{lin2024towards}.

GAI further bridges the sim-to-real gap by generating diverse hypothetical conditions such as abrupt weather shifts, biofouling accumulation, or sensor noise within the digital twin \citep{menges2024digital,ali2025digital}. These enriched simulations train autonomous ROVs and USVs to adapt control policies that are robust to real-world disturbances. Techniques like GANs and diffusion models are particularly effective in synthesizing sensor data streams (e.g., sonar, optical, or turbidity sensors) that reflect the variability of underwater environments \citep{ma2024survey, shen2025position}. Recent studies have demonstrated the use of GAI to simulate realistic underwater images and sensor patterns for training perception models \citep{ali2024foundation,raza2022towards}, while others have shown how digital twin platforms can support scenario-based mission rehearsal and online control tuning in aquaculture systems \citep{hasan2024leveraging, madusanka2023digital}. This synergy enables safer, data-driven decision-making and optimizes robotic mission planning in dynamic aquaculture environments \citep{ciuccoli2024underwater}.

\subsubsection{Multimodal Perception and Sensor Fusion}

Achieving accurate environmental perception in aquaculture robotics is inherently difficult, given the dynamic and diverse characteristics of underwater environments~\citep{aubard2025sonar}. These environments often require the integration of multiple sensing modalities, such as visual (cameras), acoustic (sonar), tactile, and environmental (e.g., pH, temperature, turbidity) data. GAI plays a transformative role in enhancing multimodal perception by enabling robust sensor fusion, adaptive reconstruction, and predictive modeling \citep{tang2023comparative,gepperth2016generative}.

Foundation models trained on large-scale multimodal datasets can learn to generate missing or noisy sensor inputs, improving perception reliability in low-visibility or high-turbidity conditions\citep{fan2024multimodal}. For example, GAI techniques such as cross-modal transformers and generative fusion networks can align asynchronous or partially missing data streams like combining degraded optical images with sonar readings to generate coherent environmental maps \citep{lee2020multimodal}. This enhances the robot’s spatial awareness, particularly for tasks like net pen inspection, biomass estimation, or debris detection \citep{huang2025survey}.

Moreover, GAI facilitates semantic-level sensor fusion by unifying structured (e.g., sonar maps) and unstructured (e.g., video frames) data into common representations that downstream models can use for high-level reasoning and decision-making \citep{wang2025large}. Recent advances have demonstrated the ability of generative multimodal models to simulate complex sensory interactions, enabling real-time decision support for precision feeding, anomaly detection, and adaptive mission planning in aquaculture systems \citep{he2023foundation, tang2023comparative, han2025multimodal}.

\subsubsection{Collaborative Multi-Agent Systems}

\begin{figure}[t]
\centering
\includegraphics[width=1\linewidth]{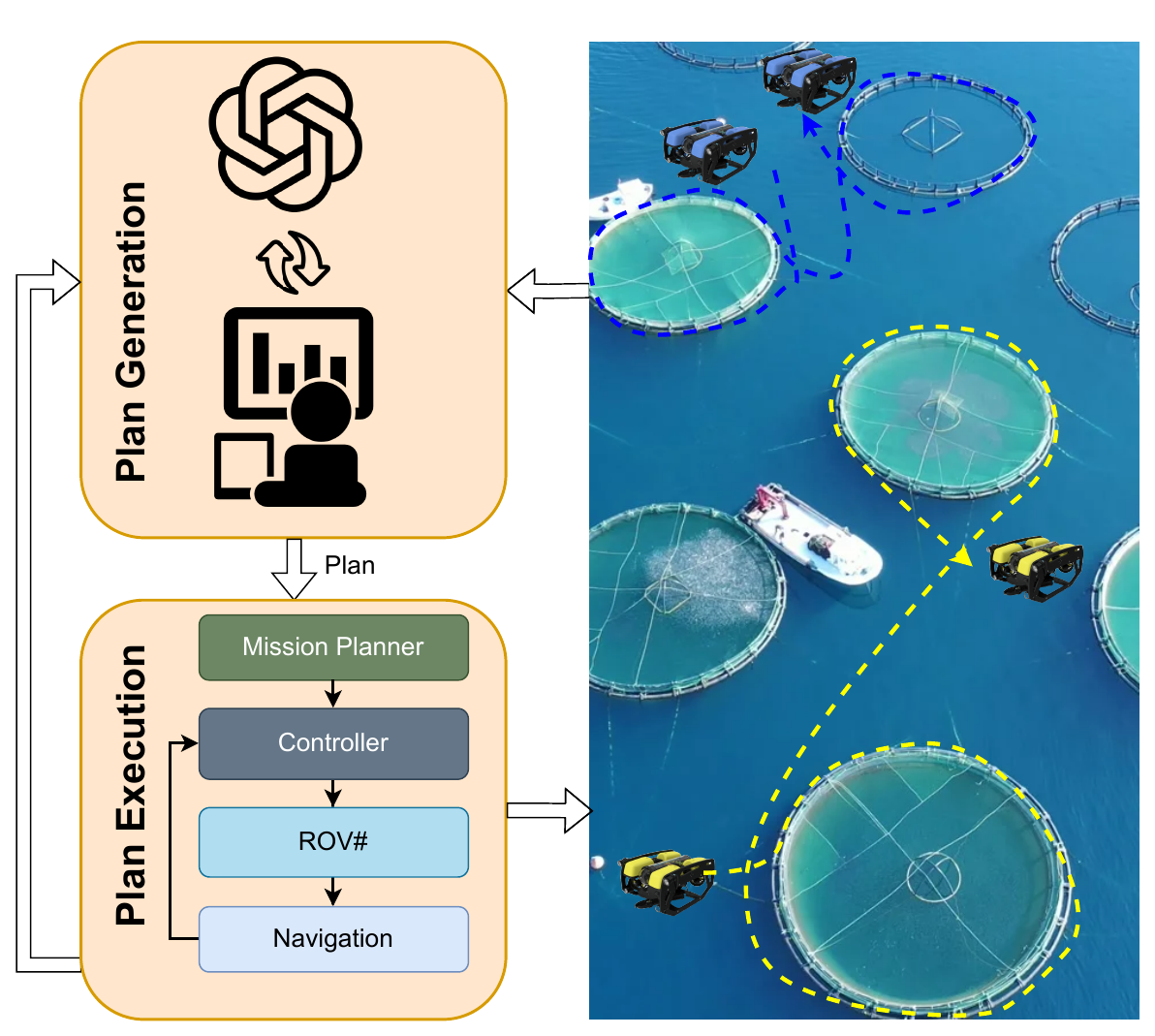}
\caption{GAI-enabled architecture for collaborative multi-agent aquaculture operations. The framework couples plan generation using foundation models with distributed execution across autonomous ROVs for real-time inspection and coordination.}
\label{fig:multirov}
\end{figure}

Collaborative multi-agent robotic systems are becoming essential in aquaculture for scalable inspection, maintenance, and monitoring \citep{song2023llm}. These systems involve fleets of ROVs, USVs, or drones working together to cover large underwater infrastructures efficiently. GAI plays a central role in enhancing the autonomy and intelligence of these systems by supporting adaptive coordination, predictive task allocation, and real-time replanning under dynamic environmental constraints \citep{wu2025generative}. An example of a GAI-enabled architecture for collaborative multi-agent operations in aquaculture sector is illustrated in Figure~\ref{fig:multirov}. 

GAI enables multi-agent systems to simulate and reason about complex inter-agent dynamics, such as communication delays, energy trade-offs, and coverage overlap \citep{xu2024multi, xiao2019marinemas}. This predictive capability allows autonomous agents to negotiate roles, anticipate conflicts, and cooperatively execute tasks with minimal supervision. For instance, generative policy models can simulate diverse operational scenarios like abrupt environmental changes or partial agent failures and suggest optimal task reallocations in real-time \citep{xu2024multi}.

Moreover, GAI empowers multi-agent planners to synthesize high-level inspection strategies and decompose them into agent-specific trajectories. Such capabilities have been used to orchestrate collaborative underwater inspections, distributed feeding schedules, and dynamic infrastructure repair missions \citep{wang2025heterogeneous}. By combining learned models with real-time environmental feedback, GAI significantly improves the resilience, adaptability, and overall effectiveness of collaborative robotic systems in aquaculture \citep{you2025generative}.

\subsection{Planning and Optimization}
\begin{figure}[t]
\centering
\includegraphics[width=1\linewidth]{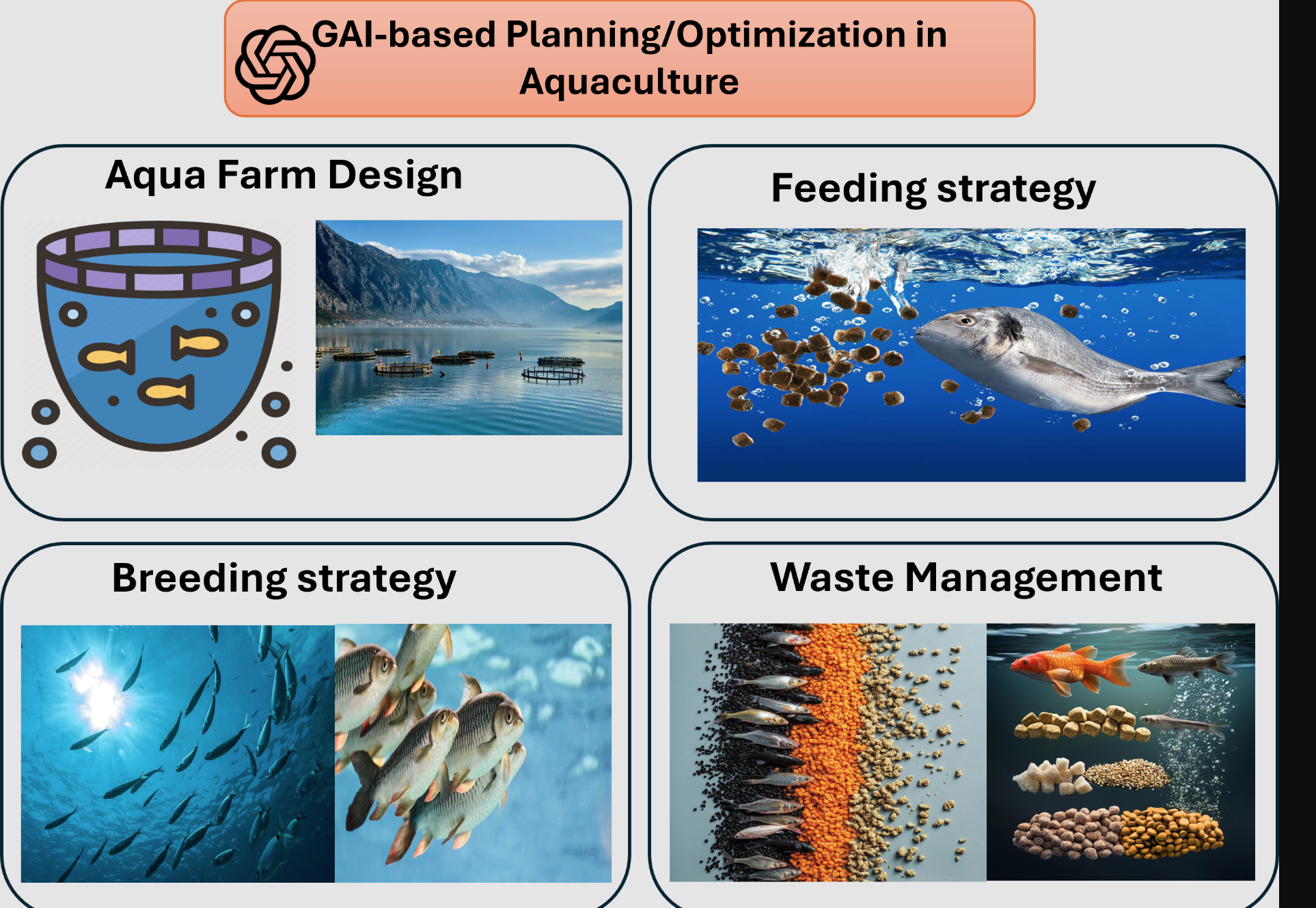}
\caption{Illustration of GAI-based planning and optimization in aquaculture, highlighting key application areas including infrastructure design, precision feeding, breeding strategies, and waste management for enhanced sustainability and operational efficiency.}
\label{fig:plan}
\end{figure}

Modern aquaculture faces growing demands for higher productivity, environmental sustainability, and cost efficiency \citep{dominguez2024review}. Achieving these goals requires sophisticated planning and optimization strategies that can respond dynamically to complex biological and environmental systems. GAI introduces powerful new capabilities for data-driven design, precise operational control, and scenario-based forecasting, transforming how farms are planned, built, and managed. Figure~\ref{fig:plan} illustrates key areas where GAI supports advanced planning and optimization workflows, including infrastructure design, precision feeding, breeding programs and waste management.

\subsubsection{Farm Design and Infrastructure Planning}

Effective aquaculture infrastructure design is pivotal for maximizing productivity, ensuring animal welfare, and achieving environmental sustainability \citep{colt2002design}. Traditionally, farm layouts and system configurations have been developed through manual engineering approaches, empirical hydrodynamic simulations, and reliance on historical environmental data. The advent of GAI, however, has introduced a transformative paradigm enabling rapid, adaptive, and highly data-driven design processes \citep{terjesen2013design}.

GAI techniques can incorporate multi-source inputs such as bathymetric profiles, real-time water quality parameters, species-specific biological constraints, and operational goals to autonomously generate and evaluate diverse aquaculture infrastructure configurations \citep{aung2025artificial}. By employing generative design approaches including diffusion models and VAEs, GAI can simulate infrastructure behavior under various environmental conditions. This allows for optimization of net pen placement to enhance hydrodynamic flow and oxygenation, strategic sensor deployment for precision monitoring, robust biosecurity zoning, and efficient spatial planning for operational access. In addition, vision-language models such as ChatGPT support the conceptual visualization of farm layouts, while AI-integrated parametric modeling tools assist in developing high-fidelity, customizable 3D infrastructure blueprints \citep{la2024exploring}.

Recent applications demonstrate the growing role of AI-augmented and generative design techniques in aquaculture infrastructure. For instance, \citep{katsidoniotaki2024digital} developed a digital twin framework to monitor and simulate real-time structural dynamics of salmon net cages in Norwegian fjords, enabling dynamic adaptation of farm layouts based on oceanic and mechanical feedback. In spatial planning, \citep{li2023autonomous} introduced an autonomous GIS system capable of algorithmically generating optimized farm siting plans using AI-driven spatial reasoning. Furthermore, reviews by \citep{wang2021intelligent} and \citep{vo2021overview} describe broader GAI applications in aquaculture infrastructure, including smart water systems, autonomous sensor placement, and predictive control loops for biosecurity and system performance enhancement. These systems highlight the capacity of GAI to process environmental, regulatory, and operational constraints into practical design recommendations.

\subsubsection{Precision Feeding and Water Quality Simulation}

Optimal feeding strategies and maintaining high water quality standards are crucial for sustainable aquaculture operations \citep{luna2022determination}. Traditionally reliant on empirical estimates and manual observations, feeding management frequently leads to inefficiencies, waste, and environmental stress. GAI methodologies have transformed this area, introducing precise predictive models and sophisticated simulation techniques that enable dynamic management of feeding schedules and water quality parameters \citep{davis2022feed,saad2024optimizing}.

GAI-driven models utilize multimodal data including real-time water temperature, dissolved oxygen levels, fish biomass, and historical feeding records to generate precise feeding schedules \citep{aung2025artificial}. Generative sequence models, such as Transformer-based architectures, have proven particularly effective in modeling fish appetite and predicting feed conversion efficiency under varying conditions, enabling significant reductions in feed wastage and operational costs \citep{wu2024identification, metin2023temporal}. For instance, applications of generative temporal modeling have demonstrated substantial feed savings by accurately predicting optimal feeding windows in salmon and tilapia farms \citep{tynchenko2024predicting,rather2024exploring}.

Moreover, GAI enables advanced simulation of water quality dynamics, incorporating transformer-based models \citep{huang2025artificial}. These generative frameworks effectively model the complex interactions between stocking densities, metabolic waste production, temperature variations, and system configurations \citep{chen2025digitizing}. The result is a detailed forecasting capability for critical parameters like ammonia concentration, pH stability, dissolved oxygen availability, and turbidity levels, empowering proactive interventions in aeration, biofilter management, and water exchange practices \citep{cui2024multimodal,spillias2025automated}. Integration of generative models with real-time sensor networks ensures adaptive and responsive control, continuously updating management decisions based on current environmental feedback. Such systems not only enhance resource efficiency but also substantially reduce the environmental footprint of aquaculture operations.

\subsubsection{Breeding Strategy Optimization}

Selective breeding programs in aquaculture traditionally require substantial investments in phenotypic and genotypic data collection, statistical modeling, and iterative experimental validation. GAI, particularly through the use of deep learning architectures and transformers, presents transformative potential to enhance and accelerate these processes \citep{killoran2017generating}.

GAI models can simulate and predict genetic inheritance and trait expression across generations by integrating heterogeneous data sources, including genomic, environmental, and phenotypic information. This capability allows aquaculturists to optimize mating strategies, improve genetic diversity, and selectively enhance economically important traits such as growth rate, feed conversion efficiency, disease resistance, and environmental adaptability, while also minimizing inbreeding depression \citep{wu2024transformer}.

Recent studies in related domains have laid the groundwork for applying GAI in aquaculture breeding. For instance, GANs have been used to generate synthetic DNA sequences to overcome data limitations in genomic prediction tasks \citep{killoran2017generating}. Such techniques could be adapted for aquaculture species with limited annotated genomic datasets, enabling the creation of high-quality synthetic single nucleotide polymorphism (SNP) profiles or haplotypes to strengthen genomic selection models. Similarly, transformer-based architectures have demonstrated success in modeling long-range genomic dependencies in crop breeding, outperforming traditional statistical models in predicting genomic estimated breeding values (GEBVs) \citep{chen2024embarrassingly}. This approach can be extended to aquaculture to capture complex genotype–phenotype relationships in species such as salmon, shrimp, and tilapia.

In addition, autoencoders and deep neural networks have shown high predictive accuracy in estimating GEBVs for aquatic species. For example, \citet{luo2024evaluation} applied deep learning models to predict disease resistance in Pacific white shrimp, achieving superior performance compared to conventional best linear unbiased prediction (BLUP) methods. Other studies have demonstrated the utility of deep models for predicting bacterial cold water disease resistance in rainbow trout \citep{vallejo2017genomic}, highlighting the potential of GAI-enhanced trait prediction in aquaculture breeding programs.

Moreover, generative models are particularly valuable when integrating multi-omics data including genomic, transcriptomic, and proteomic datasets into cohesive predictive frameworks. These models reveal intricate gene–environment interactions, supporting precision breeding strategies that are customized for specific environmental conditions and production systems~\citep{jubair2021gptransformer}. As shown in recent reviews on multi-omics integration, these approaches lead to improved prediction accuracy and deeper understanding of trait architecture in aquatic organisms \citep{suravajhala2016multi}. The application of GAI in aquaculture breeding represents a paradigm shift enabling data-driven, high-resolution selection processes that enhance genetic improvement, sustainability, and resilience across diverse aquaculture systems.

\subsubsection{Waste Management Optimization}

Effective waste management is fundamental to sustainable aquaculture, as the accumulation of uneaten feed, fish excreta, and organic residues can lead to eutrophication, hypoxia, and habitat degradation aquaculture \citep{pandey2023aquatic}. Traditional waste management strategies often rely on manual inspections, fixed cleaning intervals, and coarse-grained empirical models, limiting their adaptability and responsiveness to changing operational and environmental conditions. GAI introduces a transformative approach to waste management through advanced predictive modeling \citep{wang2024recent}.

Techniques such as recurrent neural networks (RNNs), transformer-based sequence models, and generative time-series algorithms enable dynamic forecasting of waste accumulation rates \citep{radford2021learning}. These models can process multimodal inputs, including stocking densities, feeding regimes, water flow, temperature, and growth stages to predict organic load buildup with high temporal and spatial accuracy. This forecasting ability supports the proactive control of feeding schedules, biofiltration parameters, and water recirculation operations, significantly mitigating nutrient discharge and ecological impact.

A persistent challenge in automating waste classification is the scarcity of labeled data for training vision-based systems. GANs and diffusion models address this by synthesizing diverse, high-resolution annotated datasets of sludge buildup, biofouling, and waste particulates \citep{maryleveraging}. These datasets support the development of robust image recognition pipelines that power automated monitoring of solid waste, sludge accumulation, and filter performance in Recirculating Aquaculture Systems (RAS) \citep{goodfellow2014generative}.

Critically, the emergence of GAI, particularly CLIP \citep{ho2020denoising}, BLIP \citep{li2023blip}, or GPT-4V \citep{openai2024gpt4v}, offers new capabilities for multimodal waste monitoring and decision support. These models can interpret visual waste patterns in conjunction with sensor metadata and textual logs to semantically classify waste events (e.g., excessive biofilm, filter clogging, or abnormal turbidity) and trigger context-aware interventions. By linking real-time video feeds with operational narratives, GAI enables intuitive interfaces for technicians and autonomous systems to understand and act on visual indicators of waste accumulation \citep{khiari2024enzymes}.

Moreover, GAI techniques can simulate biochemical filter dynamics under varying operational conditions, offering scenario-based optimization of nitrification, denitrification, and nutrient cycling performance \citep{schamne2024bim}. This is particularly beneficial in Integrated Multi-Trophic Aquaculture (IMTA) systems, where efficient nutrient reuse between species (e.g., fish to shellfish or macroalgae) is essential for ecological balance and profitability. Generative simulations help fine-tune loading rates, trophic ratios, and waste capture mechanisms to enhance circularity and reduce residual discharge \citep{morrissey2004waste}.

Overall, the fusion of predictive GAI models, synthetic data generation, and vision-language foundation models enables a new era of intelligent, scalable, and ecologically sound waste management in aquaculture. These tools empower operators with both autonomy and interpretability driving operational efficiency, regulatory compliance, and environmental sustainability.

\subsubsection{Aquaculture Supply Chain Optimization}

The aquaculture supply chain consists of hatcheries, grow-out farms, processing facilities, and distribution networks is inherently complex and geographically dispersed \citep{suresh2023good}. Traditional systems frequently suffer from logistics inefficiency, spoilage, and poor demand alignment. GAI and advanced foundation models offer transformative solutions through predictive, simulation, and decision-support capabilities \citep{ma2024large}.

Demand forecasting can be elevated by generative time-series architectures such as Temporal Fusion Transformers (TFTs) \citep{lim2021temporal}. Applied in agrifood systems, TFTs achieve high accuracy in predicting temporal patterns, enabling aquaculture producers to better synchronize production and inventory with market demand \citep{lin2022innovative,lim2021temporal}. Conditional GANs (cGANs) have also been utilized in supply chain contexts to generate realistic categorical demand patterns under data-scarce conditions, improving the robustness of planning systems \citep{wang2021intelligent,li2023large}.

Logistics and cold-chain integrity benefit from GAI-powered planning tools. Diffusion-based planning networks can simulate thousands of routing scenarios under constraints like cost, time, and temperature \citep{shehzad2025computer}. When combined with real-time IoT sensor data, these systems anticipate cold breaks and spoilage events, enabling dynamic rerouting and preemptive intervention \citep{lin2022innovative, romero2025vision}.

Foundation models such as LLMs and VLMs enhance interpretability and multimodal reasoning. LLMs, fine-tuned on trade agreements, regulatory policies, and historical contracts, can automatically draft negotiation terms and compliance checklists \citep{li2023large}. VLMs, trained on both imagery and text, support automated quality inspection identifying cold-chain anomalies, labeling errors, or spoilage indicators from package images or sensor logs in near real-time \citep{romero2025vision, shehzad2025computer}. Thus, GAI supports a connected, intelligent, and proactive aquaculture supply chain streamlining demand-supply alignment, enhancing cold-chain reliability, and improving stakeholder collaboration through foundation-model reasoning.

\subsection{Communication and Reporting}

\begin{figure}[t]
\centering
\includegraphics[width=1\linewidth]{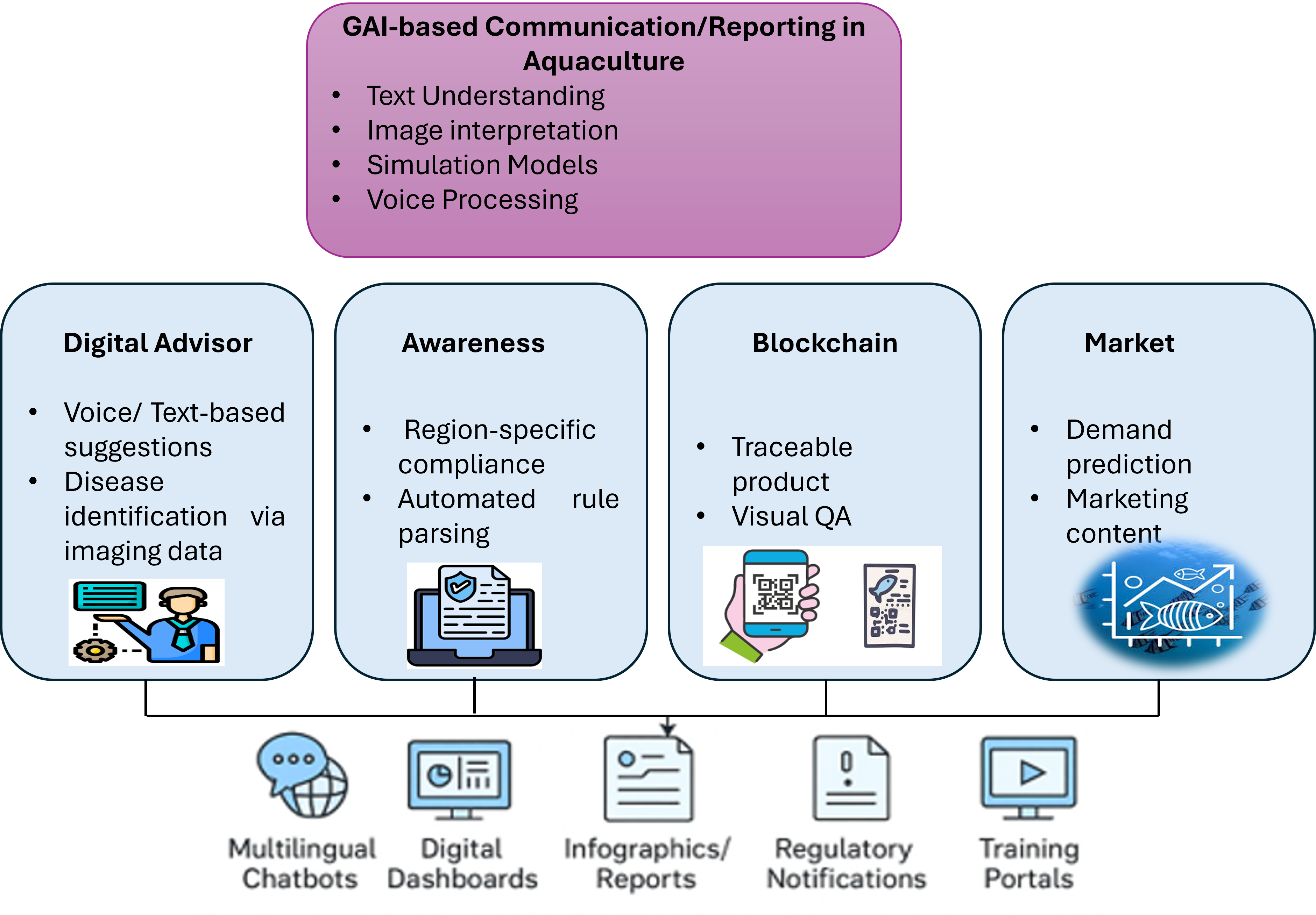}
\caption{GAI-based communication and reporting in aquaculture. Applications include personalized digital advisory, regulatory compliance support, blockchain-enabled traceability, training, and market analysis. These practices generate outputs such as multilingual chatbots, dashboards, reports, notifications, and educational portals, fostering transparent, data-driven, and scalable communication throughout the aquaculture value chain.}
\label{fig:rep}
\end{figure}

Effective communication and timely reporting are foundational pillars for the advancement of smart aquaculture systems \citep{silverthorn2025checklist}. GAI empowers these functions by enabling dynamic, context-aware, and multilingual communication channels between stakeholders, from fish farmers and regulatory bodies to consumers and supply chain actors. By leveraging natural language processing, multimodal generation, and real-time data integration, GAI enhances how information is generated, interpreted, and disseminated. This includes interactive advisory systems, regulatory updates, traceable documentation, training content, and consumer-facing narratives, all delivered through intelligent, adaptive interfaces. The result is a more transparent, informed, and responsive aquaculture ecosystem, where communication flows seamlessly across digital, regulatory, and operational layers \citep{garlock2024environmental}. Figure~\ref{fig:rep} illustrates key applications of GAI-based communication and reporting, highlighting how these technologies produce outputs such as multilingual chatbots, dashboards, reports, notifications, and educational portals to enhance transparency and stakeholder engagement across the value chain in aquaculture sector.

\subsubsection{Personalized Digital Advisors for Fish Farmers}

GAI provides a transformative approach to delivering personalized advisory services in aquaculture, bridging critical gaps created by geographic, economic, and educational limitations. By leveraging GAI, personalized digital advisors can offer real-time, contextually relevant guidance across numerous farm management activities, including feeding optimization, water quality management, disease identification, harvesting schedules, and regulatory compliance \citep{ramanan2025ai, landge2025iot}.

Unlike conventional static systems, generative models dynamically adapt their knowledge by continuously learning from real-time inputs and historical farm data, tailoring recommendations according to local environmental conditions, species-specific needs, and seasonal variability. Such adaptability ensures farmers consistently receive precise, personalized advice optimized for their unique operational contexts.

GAI further enhance user interaction through multimodal capabilities, combining textual, visual, and auditory data. For example, a farmer encountering a potential disease outbreak could provide an image of an affected fish; the generative advisor then identifies probable conditions and proposes targeted intervention measures. Similarly, voice-based interactions enable hands-free, intuitive access to critical knowledge, enhancing user experience particularly in field settings.

An illustrative example is AquaGPT \citep{aurory2024aquagpt}, a GAI-based chatbot developed that offers best-practice insights, disease detection support, and market connectivity. Another comparable initiatives include FishPal AI, offering dynamic AI-driven aquaculture recommendations, and SmartAqua AI, which integrates IoT sensor data with generative models to provide real-time water quality advice \citep{kush2025integrating}. By democratizing access to expert knowledge and providing scalable, localized, multilingual assistance, these GAI advisors significantly enhance productivity, sustainability, and operational resilience in aquaculture.

\subsubsection{Enhancing Awareness of Government Programs and Regulations}

Effective dissemination and understanding of government policies and regulations play a critical role in sustainable aquaculture development. However, many aquaculture practitioners, particularly in remote or developing regions, struggle with limited access to timely regulatory updates. GAI presents a viable solution by providing real-time, context-specific regulatory information, ensuring improved compliance and better alignment with policy objectives \citep{alexander2024responsible, ma2025redefining}.

GAI systems equipped with natural language understanding capabilities can intelligently interpret complex regulatory frameworks and provide clear, actionable guidance customized to individual farmer needs. For instance, an aquaculture operator in the UAE might interact with a GAI chatbot to obtain region-specific information on sustainable fisheries guidelines, subsidies for eco-friendly operations, or compliance protocols mandated by local authorities. These AI-powered systems continuously integrate up-to-date information from governmental databases, automatically updating their knowledge base to reflect current regulations, thus ensuring accuracy and reliability.

Real-world implementations include the UAE National Framework for Sustainable Fisheries, which utilizes AI-driven platforms to keep local aquaculture operators informed about species conservation requirements, permissible fishing practices, and resource management strategies \citep{uae2023sustainablefisheries}. Similarly, the European Union’s Common Fisheries Policy (CFP) benefits from GAI systems that provide multilingual, real-time updates regarding fisheries management, quotas, sustainability certifications, and available governmental incentives, effectively streamlining farmer compliance across the diverse regulatory landscape of the EU \citep{eu2023cfp}.

Further examples include AquaReg AI, aiding farmers in navigating complex environmental regulations and subsidy programs, and FarmAssist AI, proactively updating farmers on local regulatory developments related to water quality and fish health monitoring \citep{ramanan2025ai}. By offering dynamic content delivery, localized compliance assistance, and multilingual support, GAI significantly enhances the accessibility and practicality of regulatory guidance, fostering improved compliance and greater participation in sustainable practices across aquaculture sectors.

\subsubsection{Blockchain Integration for Traceability and Transparency}

Transparency and accountability have become fundamental in sustainable aquaculture, driven by growing consumer awareness of environmental and ethical sourcing issues \citep{iles2007making}. Integrating GAI with blockchain technology creates comprehensive, reliable traceability frameworks that not only enhance transparency but also streamline data management and regulatory compliance processes \citep{lam2016ethics,rane2023blockchain}.

Blockchain technology provides a decentralized, immutable ledger of transaction records, securing data across the aquaculture supply chain from initial stocking through harvest, processing, distribution, and retail sales \citep{leghemo2025data}. GAI complements this secure foundation by automating and enriching data capture processes. Generative models, such as GPT-4, automatically generate standardized compliance documents, export certifications, and audit reports, tailored specifically to regulatory standards across diverse markets \citep{rorvik2022blockchain,tolentino2023use}.

Additionally, generative vision models play a crucial role in data validation processes by synthetically verifying visual documentation such as product quality images or health inspections, seamlessly integrating with blockchain records \citep{rovzanec2022towards}. GAI also enables simulation of potential vulnerabilities or gaps in traceability data, proactively identifying risks or areas requiring auditing intervention \citep{vijay2023blockchain}.

Practical deployments of combined GAI-blockchain systems have already been explored in sustainable shrimp farming initiatives in Southeast Asia \citep{chase2019nfi}, such as IBM Food Trust pilots in Vietnam and Thailand \citep{shamsuzzoha2024blockchain,islam2024deep,ledgerinsights2019nfi}. These platforms utilize GAI to dynamically create multilingual labels, detailed product histories, and quality assurance visualizations derived from blockchain-verified data \citep{dhal2025leveraging,gleadall2024sustainable}.

\subsubsection{Educational Content for Aquaculture Training}

As aquaculture operations expand globally, there is a growing need for accessible, updated, and culturally relevant educational materials \citep{obiero2016knowledge}. GAI presents a groundbreaking solution to this educational challenge by dynamically generating comprehensive, adaptive, and localized content tailored specifically to diverse learning needs within the aquaculture sector \citep{brugere2023humanizing}.

Generative models, notably LLMs, have proven highly effective at automatically transforming complex technical knowledge into practical training materials \citep{fernandes2012promoting}. These models synthesize information from raw research data, field studies, and best-practice guidelines into clear, accessible manuals, guides, and instructional materials, addressing essential topics such as disease management, nutritional strategies, water quality monitoring, and emergency response planning \citep{boyd2020achieving}.

Additionally, GAI enables the creation of visual learning materials that illustrate critical biological concepts, farm infrastructure designs, and operational procedures \citep{rane2024generative}. Advanced generative simulations further enhance training experiences by providing virtual, interactive environments or digital twins, allowing learners to safely practice and refine skills in simulated real-world aquaculture scenarios \citep{li2024foundation,li2023large}.

A significant advantage of GAI in aquaculture education is its capacity for personalized adaptive learning \citep{khademi2024educational}. By analyzing learner interactions and performance, generative models identify specific knowledge gaps and dynamically adjust curricula accordingly, delivering targeted content tailored to individual learning styles and proficiency levels. Furthermore, generative models facilitate multilingual and culturally tailored content production, breaking down language barriers and expanding educational accessibility in multilingual regions such as Southeast Asia, Africa, and Latin America \citep{haese2024addressing}.

\subsubsection{AI-Driven Market Analysis}

Market intelligence and consumer preference analysis are crucial for the commercial success and sustainability of aquaculture enterprises \citep{zander2018sustainable}. GAI significantly enhances these analytical capabilities by providing advanced predictive modeling, consumer insight generation, and strategic marketing simulations based on extensive market data \citep{olawunmi2023analysing}.

GAI methodologies utilize large-scale data sets, including social media sentiment, consumer feedback, and e-commerce trends to identify emerging consumer preferences and market opportunities \citep{ccelik2021target}. Models like GPT-4 analyze textual data from consumer reviews, culinary trends, and industry reports to infer latent demand for sustainable seafood varieties, preferences for specific species, or evolving dietary trends influenced by sustainability concerns \citep{lucas2021trend,aung2025artificial}.

Generative models further contribute by simulating diverse market scenarios, such as changes in consumer demand driven by environmental factors or regulatory shifts. These sophisticated scenario analyses enable aquaculture producers to strategically adjust production volumes, optimize harvest timing, and align product offerings with forecasted market demand \citep{fernandes2024artificial}.

Moreover, personalized marketing strategies can be dynamically generated by generative models, crafting region-specific branding campaigns and consumer-centric content tailored to individual market segments \citep{rane2024generative}. A noteworthy practical implementation is seen in Alibaba’s DAMO Academy, which employs generative models for highly accurate seafood demand forecasting at city-specific levels, significantly improving inventory management and reducing product waste \citep{gladju2022applications,li2025reviews,mitra2024fishsurv}.

Despite potential challenges related to bias mitigation, data availability, and integration complexity, ongoing research continues to refine GAI methodologies for robust, ethical market analysis applications in aquaculture, promising significant future impacts on industry profitability and sustainability.

\section{Case Studies and Integration of GAI/LLMs in Marine Robotics}
\label{sec:cases}

Recent developments in GAI and LLMs have significantly transformed the capabilities of marine robotics across platforms such as AUVs, USVs, ROVs, and hybrid robotic systems. These models enable not only intelligent navigation and adaptive control but also enhance perception, multimodal reasoning, and natural language interaction. As the demand for resilient and autonomous systems in marine environments grows, GAI and LLMs are increasingly integrated into mission-critical workflows, facilitating real-time decision-making, explainable behavior, and low-power onboard intelligence.

This section synthesizes peer-reviewed case studies and implementation frameworks published between 2021 and 2025 that explore how GAI is embedded into key subsystems of marine robotics. It spans language-driven planning~\citep{yang2023oceanchat,yang2024oceanplan}, multimodal perception~\citep{samuel2024integrating,sundaravadivel2024image}, adaptive learning and control~\citep{wen2025llmauv,peng2023motioncontrol}, explainable human-robot interaction (HRI)~\citep{lin2024embodied}, and risk-aware deployment~\citep{zhang2024prompt,karim2025securing}. Emerging benchmarks such as OCEANBENCH~\citep{bi2023oceangpt} and segmentation tools like SAM2~\citep{lian2024evaluation} are also enabling reproducibility and comparative analysis across diverse marine datasets.

Table~\ref{tab:case_studies_summary} summarizes a representative collection of these systems and frameworks, highlighting their application domains, technical innovations, and relevance to future deployments in both research and commercial contexts.

\begin{table*}[ht]
\centering
\caption{Summary of GAI and LLM Integration in Marine Robotics (2021--2025)}
\label{tab:case_studies_summary}
\begin{tabular}{p{3.8cm} p{4.6cm} p{6.6cm}}
\hline
\textbf{System / Study} & \textbf{Application Domain} & \textbf{Key Contributions and Insights} \\
\hline
OceanChat~\citep{yang2023oceanchat} & AUV planning & LLM-guided closed-loop planner; HoloOcean simulation validation \\
OceanPlan~\citep{yang2024oceanplan} & Hierarchical AUV control & Mission decomposition and dynamic replanning using LLMs \\
Word2Wave~\citep{chen2024word2wave} & Natural language AUV programming & GPT + SLAM pipeline for language-driven mission control \\
FathomGPT~\citep{khanal2024fathomgpt} & Ocean data interaction & Interactive querying and taxonomic data exploration \\
OCEANGPT~\citep{bi2023oceangpt} & Ocean science QA & OceanBench benchmark and multi-agent dataset generation \\
Stonefish~\citep{grimaldi2025stonefish} & Simulation & Open-source ROS-based simulator for ML training and testing \\
Cascade Planner~\citep{nikushchenko2023cascade} & AUV path planning & Organism-inspired layered planning architecture \\
Multimodal Edge-AI~\citep{sundaravadivel2024image} & ROV/AUV coordination & LLM-vision fusion for embedded underwater control \\
Pollution Monitoring~\citep{samuel2024integrating} & Surface water sensing & LLM + Raspberry Pi + camera for pollutant detection \\
SAM2 Evaluation~\citep{lian2024evaluation} & Underwater segmentation & Prompt-based benchmarking on UIIS/USIS10K \\
COLREGs LLM~\citep{agyei2024large} & ASV navigation & Explainable LLM decision logic for maritime compliance \\
Lightweight MLLM~\citep{xi2025lightweight} & Maritime scene understanding & Multimodal sensor-text classifier for resource-constrained robots \\
Mission Explainability~\citep{coffelt2025auv} & AUV mission review & Graph + LLM system for post-mission queries and debrief \\
Bathymetry Framework~\citep{marina2024bathymetry} & Seafloor mapping & LLM-assisted spatial reasoning using bathymetric data \\
Path Planning (RLHF)~\citep{wen2025llmauv} & AUV trajectory control & RL + LLM system robust to ocean current variability \\
Swarm Control~\citep{wang2021usv, liu2024pattern} & USV coordination & Multi-agent reinforcement learning with language-guided formations \\
Trajectory Tracking~\citep{peng2023motioncontrol} & USV feedback control & Deep model-based motion tracking for underactuated vessels \\
Biomimetic AUV~\citep{anand2024bioauv} & Sustainable propulsion & AI-enabled, quiet, energy-efficient motion inspired by marine biology \\
Human-Robot Interaction~\citep{lin2024embodied} & Trustworthy autonomy & Emotional logic engine for adaptive, explainable HRI \\
Security and Privacy~\citep{zhang2024prompt} & Cyber-physical safety & Prompt-injection risks and AI risk management frameworks \\
AI Regulation~\citep{luo2024embodied} & Ethics and governance & Navigation laws and standardized protocols for marine AI \\
Commercialization Trends~\citep{fortino2024generative, fortino2025generative} & Industry impact & GAI as enabler for democratization of robotic access \\
Architecture Adaptation~\citep{qi2024advances, benjdira2025prm} & Modality-specific optimization & PRM for interfacing LLMs with sonar/acoustic sensors \\
\hline
\end{tabular}
\end{table*}

\subsection{Language-Driven Planning and Control Systems}

GAI has significantly advanced task-level autonomy in marine robotics. Systems such as OceanChat~\citep{yang2023oceanchat} and OceanPlan~\citep{yang2024oceanplan} enable LLM-guided mission interpretation, real-time replanning, and closed-loop execution for AUVs. These tools transform natural language commands into structured actions, leveraging hierarchical task decomposition and dynamic reactivity. Word2Wave~\citep{chen2024word2wave} builds on this by enabling intuitive AUV programming using GPT and SLM-based models. Furthermore, COLREGs LLM~\citep{agyei2024large} demonstrates compliance with maritime collision rules, making USVs safer and more interpretable. Beyond enabling direct control, GAI also plays a pivotal role in refining how marine robots interact with and interpret their surrounding data, leading to advancements in perception interfaces.

\subsection{Interactive Data and Perception Interfaces}

Building upon the advancements in control, LLMs are also revolutionizing ocean science workflows and enhancing robotic perception. Examples such as FathomGPT~\citep{khanal2024fathomgpt} and OCEANGPT~\citep{bi2023oceangpt} highlight how these systems support taxonomic queries, ocean chart interaction, and QA tasks via benchmarks like OCEANBENCH. Moreover, Maritime LLM-Vision~\citep{samuel2024integrating} and SAM2~\citep{lian2024evaluation} demonstrate how vision-language fusion can facilitate critical environmental monitoring. The METRICS initiative~\citep{gentili2023detection} further supports these efforts by developing tools for perception, annotation, and underwater object detection under difficult visibility conditions. To effectively develop and validate these sophisticated GAI-driven control and perception systems, robust simulation environments, sophisticated reinforcement learning techniques, and coordination strategies are indispensable.

\subsection{Simulation, Reinforcement Learning, and Coordination}

Platforms like Stonefish~\citep{grimaldi2025stonefish} and NauSim~\citep{ortiz2024nausim} serve as vital simulation environments for LLM-integrated robotic systems, offering ROS compatibility and facilitating real-time interaction with perception and control modules. Meanwhile, projects like Cascade Planner~\citep{nikushchenko2023cascade} and RLHF path planners~\citep{wen2025llmauv} combine reinforcement learning with GAI to enable dynamic navigation in complex marine environments. Furthermore, swarm control frameworks~\citep{liu2024pattern, wang2021usv} demonstrate how multi-agent reinforcement learning can support collaborative trajectory management for fleets of USVs. These technological advancements in simulation and learning are critical for developing intelligent marine robots; however, their effective and responsible deployment necessitates careful consideration of human interaction, ethical implications, and overall readiness for real-world scenarios.

\subsection{Human-Robot Interaction, Ethics, and Deployment Readiness}

The increasing sophistication of GAI in marine robotics also brings to the forefront critical considerations regarding human-robot interaction, ethical guidelines, and the practical challenges of deployment, including security and privacy. Security risks such as prompt injection have raised concerns about reliability in LLM-based control systems~\citep{zhang2024prompt}, prompting researchers to develop frameworks like Prompting Robotic Modalities (PRM)~\citep{benjdira2025prm} and align with NIST AI risk management guidelines~\citep{karim2025securing}. In parallel, HRI research has focused on making systems more transparent and adaptive. Emotional logic engines~\citep{lin2024embodied} and interactive mission explainability frameworks~\citep{coffelt2025auv} aim to build trust between humans and machines. Addressing these challenges in HRI, ethics, and security is paramount to realizing the full potential of GAI in marine robotics and unlocking its transformative industrial and societal impacts.

\subsection{Industrial and Societal Impacts}

GAI is increasingly seen as a transformative enabler in commercial marine robotics, holding significant implications for industrial growth, economic opportunities, and societal advancements. Studies by Fortino et al.~\citep{fortino2024generative, fortino2025generative} highlight the role of generative models in democratizing access to intelligent systems. Similarly, academic programs like MIR~\citep{marxer2021mir} support this trend by training interdisciplinary experts to deploy and regulate such systems. Nonetheless, broader deployment still requires robust frameworks for ethical compliance, energy efficiency, and environmental protection~\citep{luo2024embodied}. These trends underscore GAI's profound and multifaceted influence, which will be further synthesized in the following discussion regarding overarching themes, challenges, and future directions.

\subsection*{Discussion and Synthesis}

The reviewed case studies (Table~\ref{tab:case_studies_summary}) reflect a growing convergence between LLMs and multimodal marine robotics. GAI-infused systems are reshaping the control, perception, and interaction paradigms for autonomous underwater and surface vehicles, from mission decomposition and natural language programming to perceptual grounding and simulation-aided training. Notable examples include adaptive path planning under current disturbances~\citep{wen2025llmauv}, explainable mission logic~\citep{coffelt2025auv}, and maritime rule compliance through LLM reasoning~\citep{agyei2024large}.

However, despite this momentum, the field remains predominantly experimental, with limited operational validation beyond simulation environments. The deployment of GAI/LLM systems in real-world marine contexts faces several systemic bottlenecks:

\begin{itemize}
    \item \textbf{Computational constraints and energy overhead}: Most LLM-based controllers are computationally intensive and ill-suited for embedded deployment on AUVs, USVs, and ROVs due to resource constraints. Few studies (e.g., \citep{xi2025lightweight}) propose efficient alternatives, and energy-efficient inference remains a critical limitation.

    \item \textbf{Communication latency and failure modes}: High-latency or lossy acoustic communication severely limits real-time human-in-the-loop interaction and distributed LLM coordination in underwater settings. This makes cloud-dependent or multi-agent GAI systems difficult to scale in practice.

    \item \textbf{Security and adversarial vulnerability}: Prompt injection attacks and response manipulation remain underexplored but high-impact risks~\citep{zhang2024prompt}. Efforts like PRM~\citep{benjdira2025prm} and NIST-aligned risk frameworks~\citep{karim2025securing} provide initial mitigation pathways yet remain abstract and untested in maritime robotics.

    \item \textbf{Ethical and regulatory vacuum}: Although several works acknowledge the need for ethical standards~\citep{luo2024embodied}, concrete frameworks for ecological safety, liability, and cross-jurisdictional compliance (e.g., in international waters) are lacking. No current GAI system explicitly implements environmental safeguards, despite its increasing use in pollution monitoring~\citep{samuel2024integrating} or marine infrastructure inspection.

    \item \textbf{Generalization and domain robustness}: While tools like Word2Wave~\citep{chen2024word2wave} and OceanPlan~\citep{yang2024oceanplan} enable flexible mission specification, their robustness to ambiguous, multilingual, or error-prone input in harsh marine settings remains unproven. Many models rely on pre-trained weights not optimized for sonar, acoustic, or low-visibility optical data.

    \item \textbf{Reproducibility and dataset transparency}: Benchmarks such as OCEANBENCH~\citep{bi2023oceangpt} or UIIS~\citep{lian2024evaluation} are promising, yet lack comprehensive metadata standards, annotation transparency, and ecosystem support. This hinders reproducibility and hinders robust cross-system comparison.
\end{itemize}

On the positive side, a few notable trends point toward broader democratization and commercial relevance. Systems like FathomGPT~\citep{khanal2024fathomgpt} and Word2Wave have proposed user-friendly interfaces for non-experts, while the MIR program~\citep{marxer2021mir} supports cross-disciplinary talent development. However, few studies explore business models, cost-benefit trade-offs, or the role of industry-academia partnerships in scaling these innovations.

In summary, the field is advancing rapidly but unevenly. GAI integration in marine robotics is no longer a theoretical possibility but is an emerging design principle. However, to move beyond early prototypes, future efforts must (i) establish energy-efficient architectures, (ii) develop mission-specific safety mechanisms, (iii) create multilingual and multi-domain training data, and (iv) adopt transparent evaluation practices. Without addressing these foundational challenges, the potential of GAI in underwater autonomy and marine governance may remain underutilized.

\section{Limitations and Challenges}
\label{sec:limitations}

Despite the promising capabilities of GAI and LLMs in transforming aquaculture, their widespread deployment remains constrained by a range of practical, technical, and societal challenges. This section critically examines the key barriers to adoption: data availability, real-time operation, interpretability, regulation, and environmental sustainability.

\subsection{Data Scarcity in Aquaculture}

One of the most pressing limitations in applying GAI to aquaculture is the scarcity and fragmentation of high-quality, well-annotated datasets. Aquaculture environments are inherently heterogeneous, exhibiting substantial variability in species morphology, behavioral patterns, enclosure design, and environmental conditions (e.g., turbidity, salinity, lighting). Data collected across these contexts often differ significantly in sensor types, formats, and annotation granularity, making unified training and generalization across farms and regions highly challenging.

Moreover, underwater imaging data frequently suffer from quality degradation due to low visibility, biofouling, motion blur, and limited light penetration. Acoustic and sonar data, while less affected by visibility, pose challenges regarding low spatial resolution and annotation difficulty. These factors collectively hinder the development of robust, generalizable GAI models.

To overcome these constraints, future efforts must focus on cross-site data harmonization, annotation protocol standardisation, and synthetic data generation via domain-adapted diffusion models or adversarial training. Collaborative data-sharing initiatives across industry and academia while addressing data ownership and privacy concerns are essential to ensure sufficient coverage of environmental and operational variability.

\subsection{Real-Time and Computational Limitations}

Timely inference is crucial in aquaculture operations, especially for disease detection, anomaly alerts, feeding optimization, and robotic control applications. However, state-of-the-art generative models typically involve billions of parameters and require significant compute resources, making them impractical for deployment in edge settings such as floating cages, offshore farms, or remote monitoring stations.

Current marine infrastructure does not support high-throughput compute nodes or stable network access, severely limiting cloud-based inference or large-scale onboard processing. Model compression strategies, such as pruning, quantization, and knowledge distillation, offer partial solutions but often degrade performance if not carefully tuned for specific modalities (e.g., underwater images vs. sonar sequences).

Research must explore lightweight GAI architectures optimized for constrained environments alongside advancements in on-device learning, federated model updates, and task-specific transformer variants. Hardware acceleration through AI-specific chips (e.g., TPUs, EdgeTPUs) could enable real-time decision support without continuous cloud connectivity.

\subsection{Trust, Transparency, and Resistance}

A persistent barrier to GAI adoption in aquaculture lies in the perceived opacity of model decisions. Stakeholders, especially farmers, technicians, and regulators, often require traceable and intuitive explanations for AI-generated insights, particularly in high-stakes tasks such as mortality prediction, disease classification, or biomass estimation.

Even if statistically accurate, black-box outputs are insufficient in domains where operational consequences carry ecological, regulatory, or economic weight. The lack of interpretability undermines user confidence, fosters resistance to automation, and can hinder regulatory approval.

To address these challenges, GAI systems must incorporate explainability techniques like attention heatmaps, counterfactual examples, and interpretable representations in latent space. More importantly, explainability should be aligned with domain semantics; for example, showing temporal behavior changes rather than abstract feature maps may be more relevant to aquaculture operators.

\subsection{Regulatory and Ethical Considerations}

As GAI systems increasingly use autonomous decisions in aquaculture, they must adhere to evolving regulatory standards and ethical expectations. Using sensitive farm-level data, e.g., real-time disease incidents, proprietary feeding patterns, or behavioural logs, raises important questions about privacy, consent, and data stewardship.

Moreover, GAI-generated outputs influencing resource usage, treatment decisions, or animal handling may lead to unintended ecological or welfare consequences. Establishing liability is particularly difficult in such scenarios when decisions emerge from non-deterministic, learning-based systems.

To address these issues, transparent frameworks must be co-developed by policymakers, technologists, and aquaculture stakeholders. These should include data governance protocols, auditing mechanisms, and clear delineation of accountability across AI-assisted workflows. Regulatory sandboxing and ethics-by-design principles can also help evaluate GAI deployment risks before full-scale integration.

\subsection{Sustainable AI Infrastructure}

Although GAI may offer long-term ecological benefits through optimized feeding, reduced waste, and early disease control, its underlying computational infrastructure is energy-intensive. Training and deploying large generative models typically involve high-power GPU clusters and centralized cloud facilities, contributing significantly to greenhouse gas emissions.

This paradox, where a technology intended to improve sustainability may be environmentally costly, must be addressed through lifecycle assessments, low-energy model design, and carbon-conscious deployment strategies. Using renewable-powered computing facilities, promoting edge inference over cloud reliance, and dynamically adjusting model size based on task sensitivity are promising directions.

Ultimately, the environmental benefits of deploying GAI in aquaculture must demonstrably outweigh its carbon footprint, and future work should aim to quantify this trade-off as part of sustainability evaluations.

\section{Future Research Directions}\label{sec:future}

As aquaculture systems become increasingly data-driven, the integration of GAI presents great potential. Future research must address emerging challenges and explore novel capabilities across perception, autonomy, and collaboration to fully utilise this technology. This section outlines key directions to shape the next generation of intelligent aquaculture systems, emphasizing multimodal integration, domain adaptation, and standardized evaluation.

\subsection{Multimodal GAI Integration}

One promising direction involves the development of multimodal GAI systems capable of fusing video, sonar, textual metadata, and environmental sensor streams. Such models can enable a more holistic understanding of aquaculture environments, improving tasks like fish behaviour analysis, disease detection, and net integrity assessment. Advancements in cross-modal fusion strategies unified latent representations, and alignment techniques will be essential to ensure these models effectively operate under noisy, heterogeneous, and dynamic underwater conditions.

\subsection{Federated GAI for Collaboration}

To address data privacy concerns and promote cross-site model collaboration, federated learning represents a valuable pathway for future exploration. In this paradigm, individual farms can train models locally while contributing to a global GAI model without sharing raw data. Key research challenges include designing robust federated algorithms for aquaculture datasets, developing secure aggregation protocols, and ensuring real-time learning performance in bandwidth-limited and hardware-constrained environments.

\subsection{Domain-Specific Pretraining}

Despite the success of general-purpose foundation models, their performance often degrades when applied to underwater robotics and aquaculture-specific tasks. Domain-specific pretraining using custom underwater imagery, sonar, and operational logs datasets can improve model robustness and contextual understanding. Future research should focus on assembling large-scale aquaculture datasets, defining practical pretraining objectives, and benchmarking domain-specialized models against their generalist counterparts to quantify improvements in performance and adaptability.

\subsection{GAI for Robotic Autonomy}

Most current GAI applications in aquaculture remain focused on perception or monitoring. Extending these systems toward underwater manipulation for tasks such as targeted net repair, precision harvesting, or invasive species removal, offers substantial operational benefits. Research into generative reinforcement learning, sim-to-real transfer, and adaptive control policies will enable GAI-driven autonomous interventions in unstructured, high-variability underwater settings.

\subsection{GAI Benchmark Development}

The lack of standardized benchmarks and reproducible evaluation protocols hinders progress in aquaculture-focused GAI. Establishing open-access datasets and clearly defined tasks ranging from low-visibility object detection to multimodal reasoning will help unify evaluation and accelerate progress. Future initiatives should prioritize benchmark development that reflects real-world environmental variability, annotation inconsistencies, and operational constraints commonly encountered in aquaculture deployments.

\section{Conclusion}\label{sec:conclusion}

GAI is emerging as a transformative force in aquaculture, offering unprecedented opportunities to enhance operational efficiency, ecological resilience, and decision-making precision. This review has demonstrated that GAI encompassing foundation models, LLMs, and generative multimodal systems have shown applicability across many aquaculture domains, including infrastructure inspection, precision feeding, health diagnostics, underwater robotics, and environmental monitoring. These technologies enable context-aware, proactive, and adaptive management strategies, supporting the transition toward intelligent and sustainable aquaculture systems.
However, realizing the full potential of GAI in this domain requires addressing several persistent challenges. Data scarcity and heterogeneity, real-time deployment constraints, limited interpretability, and the need for ethical and regulatory safeguards remain central barriers to adoption. Tackling these limitations will demand sustained, interdisciplinary collaboration, bringing together expertise from aquaculture science, machine learning, marine robotics, environmental engineering, and policy governance.
This review provides a consolidated foundation for advancing GAI research and deployment in aquaculture. By synthesizing the state of the art, critically evaluating current limitations, and outlining actionable future directions, including domain-specific model development, federated learning, underwater manipulation, and benchmarking, this work aims to guide both academic inquiry and industrial innovation. Ultimately, GAI's responsible integration into aquaculture promises to enable scalable, explainable, and environmentally conscious systems for the next generation of ocean-based food production.

\section*{Acknowledgement}
\noindent This work is supported by the Khalifa University under Award No. RC1-2018-KUCARS-8474000136, CIRA-2021-085, MBZIRC-8434000194, KU-BIT-Joint-Lab-8434000534 and KU-Stanford :8474000605.

\section*{Decleration}
During the preparation of this work the author(s) used ChatGPT-4 in order to improve language and readability. After using this tool/service, the author(s) reviewed and edited the content as needed and take(s) full responsibility for the content of the publication.

\printcredits

\bibliographystyle{model1-num-names}

\bibliography{cas-refs}


\end{document}